\documentclass[runningheads]{llncs}

\usepackage{eccv}

\usepackage{eccvabbrv}

\usepackage{graphicx}
\usepackage{booktabs}

\usepackage[accsupp]{axessibility} 

\usepackage{multirow}
\usepackage{float}

\usepackage{hyperref}

\begin{document}

\title{Argus: Metric Panoramic 3D Reconstruction for Indoor Scenes}


\author{Xi Li\inst{1} \and Linyuan Li\inst{1} \and Yan Wu\inst{1} \and Tong Rao\inst{1} \and Kai Zhang\inst{1} \and Xinchen Hui\inst{1} \and Cihui Pan\inst{1,2}\thanks{Corresponding author: \texttt{pancihui001@realsee.com}}}

\authorrunning{Xi Li et al.}

\institute{Realsee, China \and Quanzhou University of Information Engineering, China}

\maketitle

\begin{figure}[H]
	\centering
	\includegraphics[width=1\linewidth]{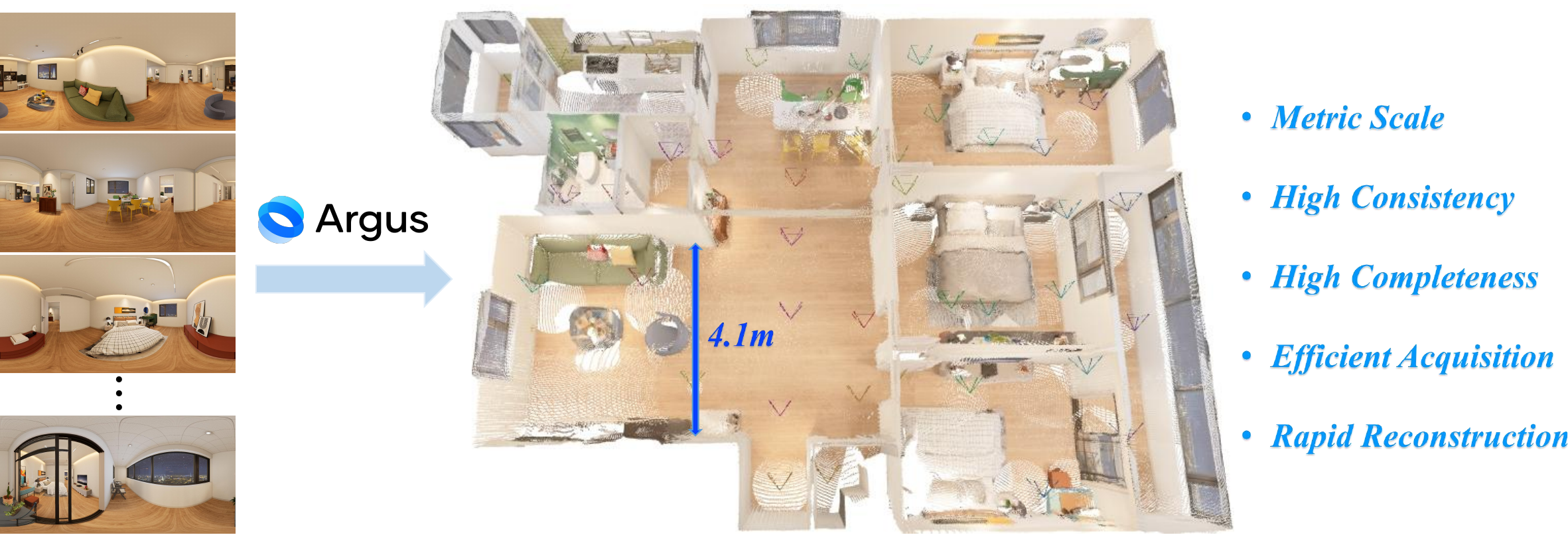}
	\caption{
		Argus is a feed-forward 3D reconstruction network trained on our large-scale indoor panoramic 3D dataset Realsee3D.
		Given acquired sparse multi-view panoramic images, it rapidly reconstructs complete, consistent, and metric-scale 3D scenes.
	}
	\label{fig:teaser}
\end{figure}

\begin{abstract}
	Metric feed-forward 3D reconstruction for panoramic data remains under-explored due to the lack of large-scale panoramic RGB-D training data.
	We present Realsee3D, a hybrid dataset of 10K indoor scenes (1K real, 9K synthetic) with 299K panoramic viewpoints and precise metric annotations,
	and Argus, a feed-forward network trained on it for metric panoramic 3D reconstruction.
	In the sparse unordered capture setting of Realsee3D,
	a poorly chosen coordinate anchor can cause global pose drift.
	Argus addresses this with a learned covisibility module that selects the geometrically optimal reference view to anchor the metric world frame.
	To further improve multi-task learning,
	we decompose the bidirectional pixel-to-world mapping into interpretable sub-steps with per-step supervision and cross-coordinate joint constraints,
	reinforcing geometric consistency across prediction branches.
	On the Realsee3D benchmark,
	Argus achieves state-of-the-art metric performance in camera pose estimation, depth estimation, and point cloud reconstruction.
	Project page: \url{https://argus-paper.realsee.ai}.
\end{abstract}

\section{Introduction}
\label{sec:intro}
Image-based 3D reconstruction is a fundamental problem in computer vision,
widely applied to robotics~\cite{hu2023toward, keetha2024splatam, alama2025rayfronts},
AR/VR~\cite{dai2017bundlefusion, Jiang2024VR-GS},
autonomous driving~\cite{fei2024driv3r, ma2019accurate},
visual localization~\cite{hausler2021patch, taira2018inloc, deng2025sail}, and other fields~\cite{li2022multi}.
Recently,
Transformer-based feed-forward 3D reconstruction has achieved breakthroughs in perspective image scenarios~\cite{dust3r_cvpr24,wang2025vggt, lin2025depth}.
Extending such methods to panoramic imagery is particularly appealing:
panoramic cameras are increasingly adopted for real-world indoor 3D capture~\cite{ai2025survey},
as a single panorama covers the full field of view and provides substantial cross-view overlap,
aligning well with the sparse-view feed-forward paradigm by enabling rapid scene-level reconstruction from only a few captures.
Despite its high practical value,
metric feed-forward 3D reconstruction tailored for panoramic data remains largely unaddressed.
Mainstream feed-forward models such as VGGT~\cite{wang2025vggt} possess multi-view geometric reasoning capabilities,
but are trained on perspective RGB-D datasets and suffer severe performance degradation when directly applied to panoramic data.
A key bottleneck lies in the scarcity of large-scale, high-quality panoramic 3D training data with accurate metric annotations.
To fill this gap,
this paper presents Realsee3D,
a large-scale indoor multi-view panoramic RGB-D dataset with room-level coverage and precise metric annotations,
and Argus,
a feed-forward network trained on Realsee3D that integrates learnable covisibility-guided reference selection with geometric factorization supervision for metric panoramic 3D reconstruction.

The Realsee3D dataset aligns with real-world panoramic capture scenarios,
where images are sparsely sampled and provided as an unordered set.
In this practical setting,
existing feed-forward reconstruction methods~\cite{wang2025vggt,keetha2025mapanything} anchor the global coordinate system to a pre-defined heuristic reference view.
When the anchor is an isolated or boundary viewpoint,
insufficient cross-view covisibility degrades geometric constraints,
causing severe global pose drift and spatially inconsistent reconstruction.
$\pi^3$~\cite{wang2025pi} adopts permutation-equivariant architectures to address permutation sensitivity,
but the tightly coupled $N\times N$ pairwise constraints require a fixed-reference branch to stabilize training from scratch.
Traditional methods~\cite{schonberger2016structure} typically rely on hand-designed heuristics to select a robust world coordinate reference,
e.g., the highest-feature view or the pair with most matches.
Inspired by these methods,
we propose a learnable covisibility module as an alternative.
Instead of a fixed heuristic anchor,
our module predicts global inter-view connectivity to dynamically select the optimal reference frame,
suppressing drift from degenerate viewpoints and improving cross-view consistency.
The pipeline retains approximate permutation equivariance,
while inheriting the smooth optimization landscape of reference-frame-oriented metric learning.

Beyond reference selection, effective multi-task supervision is critical for reconstruction quality.
Recent works~\cite{dust3r_cvpr24, wang2025vggt, wang2025moge, wang2025moge2} show that joint supervision over point maps, poses, and depth yields strong geometric synergies.
Building on this insight,
we propose an overcomplete geometric factorization supervision strategy that decomposes pixel-to-world transformations under the panoramic model into interpretable sub-steps,
each supervised individually and jointly across coordinate frames.
This lowers optimization difficulty while enhancing multi-task synergy.
We further observe empirically that sufficient high-quality metric data, combined with expressive modeling, suppresses ERP distortion and boundary artifacts without task-specific designs.

In summary, the main contributions of this work include:
\begin{itemize}
	\item A large-scale indoor 3D dataset Realsee3D, and a data-driven feed-forward metric panoramic 3D reconstruction network Argus.
	\item A covisibility-based reference view learning method that anchors the metric coordinate system and enhances reconstruction robustness to pose drift.
	\item An overcomplete geometric factorization supervision strategy that decomposes pixel-to-world transformations into supervised sub-steps with cross-coordinate consistency constraints, boosting multi-task synergy.
	\item A new panoramic 3D reconstruction benchmark on Realsee3D, on which Argus achieves best overall performance across all evaluated tasks.
\end{itemize}

\section{Related Work}
\label{sec:relatedwork}
\subsection{Traditional 3D Reconstruction}
Traditional 3D reconstruction relies on optimization-based pipelines with explicit geometric modeling.
Classical Structure-from-Motion (SfM) pipelines~\cite{agarwal2009building,schonberger2016structure, moulon2016openmvg} recover camera poses and sparse point clouds through a multi-stage process:
inter-view feature matching~\cite{lowe2004distinctive, rublee2011orb},
robust two-view geometry estimation with RANSAC~\cite{barath2018graph} and other estimators~\cite{bian2017gms, sun2020acne},
view graph construction~\cite{mur2015orb, campos2021orb}, and global bundle adjustment refinement~\cite{triggs1999bundle, hartley2003multiple}.
Given the recovered poses,
Multi-View Stereo (MVS) densifies geometry via multi-view photometric reasoning,
spanning classical pipelines~\cite{seitz2006comparison, goesele2006multi, furukawa2015multi, schonberger2016pixelwise} and learning-based approaches~\cite{yao2018mvsnet, huang2018deepmvs, wang2023adaptive}.
Recent efforts increasingly integrate learned components into the SfM/MVS pipeline,
including keypoint detectors~\cite{detone2018superpoint, zhao2022alike, he2024detector},
learned matchers~\cite{sarlin2020superglue, lindenberger2023lightglue},
detector-free semi-dense~\cite{sun2021loftr,wang2024eloftr, Li_2025_ICCV} and dense matching~\cite{edstedt2023dkm, edstedt2024roma},
monocular depth priors~\cite{depthanything, lin2025depth, wang2025moge}, and end-to-end frameworks~\cite{duisterhof2025mast3r, jang2025pow3r, yang2025fast3r}.
While these hybrid approaches achieve high accuracy,
they still inherit multi-stage iterative optimization~\cite{wang2024vggsfm, zhao2025diffusionsfm, jung2025im360}, incurring prohibitive cost for large image sets.
\subsection{Feed-Forward 3D Reconstruction}
Feed-forward methods~\cite{wang2025vggt, liu2025worldmirror, chen2025ttt3r} eliminate iterative optimization and directly predict scene geometry and camera motion from input images.
Transformer architectures~\cite{vaswani2017attention, dao2022flashattention, darcet2023vision} have emerged as the dominant backbone for this task.
DUSt3R~\cite{dust3r_cvpr24} redefines feed-forward geometry prediction as dense point map regression and achieves robust pairwise reconstruction.
VGGT~\cite{wang2025vggt} introduces alternating attention and multi-task supervision for multi-view geometry learning.
MapAnything~\cite{keetha2025mapanything} expands this paradigm to universal metric reconstruction with support for multi-modal prior inputs.
However, most multi-view feed-forward methods~\cite{wang2025vggt, zhang2025flare, streamVGGT} remain highly sensitive to input view order due to their implicit coordinate anchoring to a selected reference view.
$\pi^3$~\cite{wang2025pi} eases this sensitivity via a permutation-equivariant design for unordered input robustness.
More recently, VGGT-$\Omega$~\cite{wang2026vggt} presents the first empirical validation of the scaling law for feed-forward 3D reconstruction.
Despite these rapid advances,
existing feed-forward methods are developed and evaluated exclusively on perspective imagery,
leaving panoramic scenes largely unexplored.
Our work bridges this gap by introducing a feed-forward model for metric panoramic reconstruction.
\subsection{Panoramic 3D Reconstruction and Datasets}
The rapid progress of feed-forward methods~\cite{wang2025vggt, shen2025fastvggt, feng2025quantized} underscores that high-quality 3D datasets~\cite{reizenstein2021common, yao2020blendedmvs, ling2024dl3dv, cabon2020virtual} are the cornerstone of robust feed-forward 3D reconstruction~\cite{MegaDepthLi18, dai2017scannet, deitke2023objaverse}.
However, most large-scale indoor RGB-D datasets~\cite{antequera2020mapillary, szot2021habitat, straub2019replica, zheng2023pointodyssey} are built on perspective projection~\cite{greff2022kubric, xia2024rgbd, roberts2021hypersim, pan2023aria},
leaving a critical gap in dedicated training data and standardized benchmarks for panoramic metric reconstruction.
Existing panoramic 3D datasets such as Matterport3D~\cite{chang2017matterport3d} and Stanford2D3D~\cite{armeni2017joint} primarily serve as evaluation benchmarks for tasks like monocular depth estimation~\cite{lin2025dap, Guo2025DepthAnyCamera, piccinelli2025unik3d} and semantic segmentation~\cite{zhong2025omnisam},
but their limited scale and annotation schemes cannot support learning-based multi-view metric reconstruction.
Consequently, panoramic monocular depth estimation methods~\cite{cao2025panda, jiang2025depth, li2025depth} typically resort to transfer learning from foundation models pre-trained on large-scale perspective images such as DepthAnything V2~\cite{yang2024da2}.
To advance this under-explored field,
we introduce Realsee3D, a large-scale multi-view panoramic RGB-D dataset with precise metric annotations,
enabling training and evaluation of feed-forward panoramic reconstruction models and flexible extension to downstream tasks including depth completion~\cite{yan2022multi}, novel view synthesis~\cite{mildenhall2021nerf, kerbl3Dgaussians, jiang2025anysplat}, and embodied intelligence~\cite{zheng2025panorama}, among others.
\section{Realsee3D Dataset}
\label{sec:dataset}
\subsection{Dataset Overview and Construction}
We introduce Realsee3D\footnote{Dataset available at \url{https://dataset.realsee.ai}.}, a large-scale high-resolution panoramic dataset with metric annotations combining photorealistic real-world captures and diverse synthetic scenes,
containing 10,000 indoor scenes, 95,962 rooms, 299,073 panoramic viewpoints, and two complementary subsets detailed below.
\par\noindent\textbf{Real Subset.}
This subset consists of 1,000 real-world residential scenes, covering 9,483 rooms and 24,263 panoramic RGB-D viewpoints.
We acquire the data using tripod-mounted Realsee Galois 3D LiDAR cameras with nearly co-centered RGB and LiDAR sensors,
which greatly simplifies image stitching and ensures geometric consistency between appearance and depth measurements.
At each capture position,
the camera rotates to five preset orientations.
To preserve photometric fidelity,
we stitch images directly in the RAW domain and adopt HDR multi-exposure stacking to handle complex indoor lighting.
After on-site initial pose estimation,
we perform global registration and pose refinement,
which jointly optimizes visual and 3D features across all scans via covisibility,
with manual corrections for accurate trajectory alignment when necessary.
The final output includes high-resolution ERP RGB images,
sparse metric depth maps,
and precise camera poses,
faithfully capturing real-world illumination, textures, and object distributions.
\par\noindent\textbf{Synthetic Subset.}
To further scale the dataset and enrich scene diversity,
we construct a synthetic subset of 9,000 procedurally generated scenes,
covering 86,479 rooms and 274,810 panoramic viewpoints.
All scenes are generated through a multi-stage pipeline grounded in real-world floorplan distributions.
We adopt a hybrid rule- and learning-based discrete optimization framework that integrates real-scene layout priors and expert design knowledge to produce plausible scene configurations.
We leverage a proprietary 3D asset library containing over 100,000 high-fidelity models across 200+ semantic categories.
These models preserve fine-grained geometry and use physically-based rendering (PBR) materials,
effectively reducing the synthetic-real domain gap.
After layout generation,
a graph-based viewpoint selection algorithm determines optimal capture positions by simulating real-world roaming trajectories,
ensuring full and high-quality coverage of the 3D environment and yielding dense multi-view observations per room.
Finally, we render scenes at these viewpoints using a customized Unreal Engine 5 pipeline with hardware-accelerated ray tracing,
which simulates complex light transport including multi-bounce global illumination and ambient occlusion.
Each scene is rendered under five distinct illumination schemes (Warm Day, Cold Day, Natural Day, Warm Night, Cold Night) to maximize intra-scene diversity.
This pipeline produces dense, high-fidelity RGB-D data and precise camera poses,
providing perfect geometric ground-truth free from sensor noise in real scans.
The hybrid design of Realsee3D ensures both photorealistic diversity and accurate geometric annotations,
making it substantially larger and more diverse than existing panoramic datasets for dense 3D reconstruction.
\begin{table}[tb]
	\centering
	\caption{\textbf{Comparison with Existing Indoor Panoramic Datasets}.
		We distinguish unique viewpoints (spatial capture locations) from total images for accurate comparison.
		Realsee3D offers an unprecedented scale.}
	\label{tab:dataset_comparison}
	\resizebox{\linewidth}{!}{
		\begin{tabular}{lcccccc}
			\toprule
			Dataset                                   & Type               & Scenes          & Rooms           & Unique Viewpoints & Images           & Modalities           \\
			\midrule
			Stanford2D3D~\cite{armeni2017joint}       & Real               & 6               & 270             & 1,413             & 1,413            & RGB-D, Pose          \\
			Matterport3D~\cite{chang2017matterport3d} & Real               & 90              & 2,056           & 10,800            & 10,800           & RGB-D, Pose          \\
			ZInD~\cite{cruz2021zillow}                & Real (Unfurnished) & 1,524           & --              & 71,474            & 71,474           & RGB, Pose            \\
			Structured3D~\cite{zheng2020structured3d} & Synthetic          & 3,500           & 21,835          & 21,835            & 196,515          & RGB-D, Pose          \\
			\midrule
			\textbf{Realsee3D (Real)}                 & Real (Furnished)   & 1,000           & 9,483           & 24,263            & 24,263           & RGB-D, Pose          \\
			\textbf{Realsee3D (Synthetic)}            & Synthetic          & 9,000           & 86,479          & 274,810           & 274,810          & RGB-D, Pose          \\
			\textbf{Realsee3D (Total)}                & \textbf{Hybrid}    & \textbf{10,000} & \textbf{95,962} & \textbf{299,073}  & \textbf{299,073} & \textbf{RGB-D, Pose} \\
			\bottomrule
		\end{tabular}
	}
\end{table}
\subsection{Comparison with Existing Datasets}
As shown in Table~\ref{tab:dataset_comparison},
we compare Realsee3D against several representative indoor panoramic datasets.
While Stanford2D3D~\cite{armeni2017joint} and Matterport3D~\cite{chang2017matterport3d} lay the foundation for indoor scene understanding with high-quality real-world panoramic data,
they are limited in scene count.
ZInD~\cite{cruz2021zillow} provides massive real panoramas, yet focuses on unfurnished scenes and lacks dense metric depth.
Structured3D~\cite{zheng2020structured3d} presents large-scale synthetic data with fine annotations,
yet its 196,515 images are captured from only 21,835 unique spatial viewpoints (one per room) under varying lighting and furniture configurations.
Realsee3D complements these efforts as a large-scale hybrid dataset with dense spatial coverage,
establishing a robust new benchmark for metric 3D reconstruction.

\section{Method}
\label{sec:Method}
\begin{figure}[tb]
	\centering
	\includegraphics[width=1\linewidth]{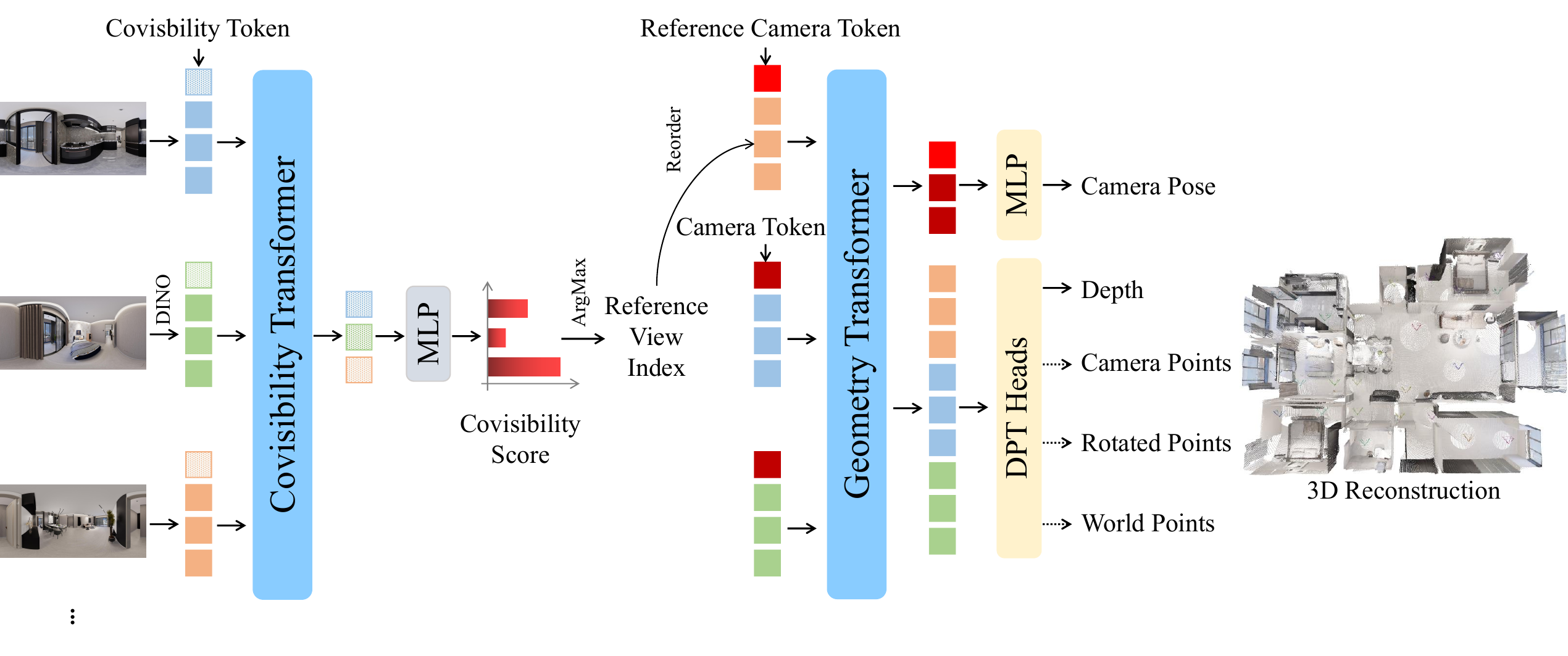}

	\caption{
		\textbf{Overview of Argus.}
		We first select the optimal reference frame via a Covisibility Transformer,
		then aggregate multi-view geometric features through a reference-based Geometry Transformer,
		followed by multiple prediction branches for intermediate geometric decomposition representations that facilitate multi-task learning.
	}
	\label{fig:network}
\end{figure}
\subsection{Overview of Argus}
An overview of our network is shown in \cref{fig:network}.
Given unordered panoramas,
our model first leverages DINOv2~\cite{oquab2023dinov2} to generate patch tokens.
Covisibility tokens are added to per-view tokens and fed into a lightweight Covisibility Transformer with $L_c=2$ alternating attention layers~\cite{wang2025vggt}
(self-attention within each view alternated with self-attention across views) and then fed into an MLP layer to predict covisibility scores.
The highest-scoring view is chosen as the reference frame with a dedicated reference camera token,
while other views use standard camera tokens.
The aggregated token sequence is then processed by a Geometry Transformer with $L_g=24$ alternating attention layers.
Finally, camera poses are regressed from camera tokens by an MLP layer that outputs a 9-dimensional vector per view:
a 3D translation, a 4D quaternion rotation, and 2 confidence scores (one for rotation, one for translation),
and depth and point maps across coordinate systems are predicted from image tokens via distinct DPT~\cite{ranftl2021vision} heads.
Each DPT head outputs an extra channel for pixel-wise confidence.
The depth head is activated with $\exp(\cdot)$ to ensure positivity,
point map heads use an inverse log transform $f(x)=\text{sign}(x)\cdot(\exp(|x|)-1)$ to handle the wide dynamic range of 3D coordinates,
where $x$ denotes the raw head output,
and all confidence values are activated via $1+\exp(\cdot)$ to guarantee a lower bound of 1.
\subsection{Learning to Select Reference View}
To enable the model to learn optimal reference frames, we construct the supervision labels as follows.
Let the image set be $\mathbb{I} = \{I_1, I_2, \dots, I_N\}$,
where $N$ is the total number of images.
We precompute the covisibility scores of all image pairs based on pose and depth and then construct a global covisibility matrix $\mathbf{M}$.
Afterwards, Dijkstra's algorithm~\cite{dijkstra1959note} is adopted to select the optimal reference view $\mathcal{I}$.
More details are provided in the supplementary materials.
The loss is computed by measuring the discrepancy between predicted covisibility logits $\hat{\mathbf{C}} \in \mathbb{R}^N$
and the one-hot ground-truth $\mathbf{C} \in \mathbb{R}^N$.
The ground-truth sets $\mathbf{C}_{\mathcal{I}}=1$ for the optimal view $\mathcal{I}$ with maximum global covisibility and $0$ otherwise ($\mathcal{I} \in \{1,\ldots,N\}$).
We adopt binary cross-entropy (BCE) loss for supervision during training.
During inference,
we select the reference frame $\hat{\mathcal{I}}$ with the highest score via the $\arg\max$ operation,
which provides approximate permutation equivariance since argmax is independent of input ordering.
For implementation simplicity,
we swap the tokens of this view with those of a fixed reference frame (the first frame) to achieve an equivalent effect.
\label{sec:trans_decomp}
\begin{figure}[tb]
	\centering
	\includegraphics[width=1\linewidth]{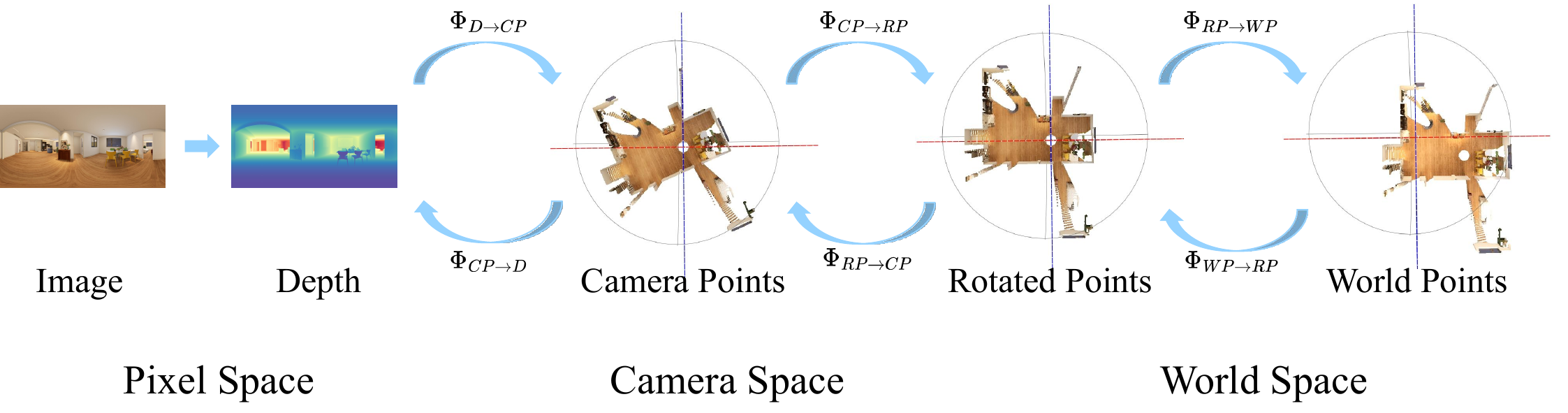}
	\caption{
		\textbf{Pixel-to-World Bidirectional Transformations for a Single View.}
	}
	\label{fig:decomp}
\end{figure}
\subsection{Panoramic Geometric Factorization}
Directly regressing world-coordinate point maps from images conflates multiple geometric transformations into a single prediction step,
which increases optimization difficulty and limits cross-task supervisory signals.
To address this, we predict each intermediate representation independently using distinct DPT~\cite{ranftl2021vision} heads
and supervise intermediate representations of forward and inverse pixel-to-world coordinate transformations in panoramic projections
via both independent and joint cross-coordinate geometric constraints (refer to \cref{sec:loss}).

For $W\times H$ ERP panoramas,
2D pixel coordinates $(u,v)$ map directly to spherical latitude $\theta = \left( \frac{v}{H} - 0.5 \right) \cdot \pi$ and longitude $\phi = \left( \frac{u}{W} - 0.5 \right) \cdot 2\pi$,
enabling computation of 3D point clouds on the unit sphere as:
\begin{equation}
	P_{u} = \begin{bmatrix}
		\cos\left(\theta\right) \cdot \sin\left(\phi\right) \\
		\sin\left(\theta\right)                             \\
		\cos\left(\theta\right) \cdot \cos\left(\phi\right)
	\end{bmatrix}
	\label{eq:sphere}
\end{equation}
All forward and inverse pixel-to-world transformations $\Phi$ for a single view are shown in \cref{fig:decomp},
which can be decomposed into the following:
\begin{equation}
	\begin{cases}
		\Phi_{D\to CP}: \, P_{c} = D\odot P_{u},            \\
		\Phi_{CP\to RP}: \, P_{r} = R P_{c},                \\
		\Phi_{RP\to WP}: \, P_{w} = P_{r} + \boldsymbol{t}, \\
		\Phi_{WP\to RP}: \, P_{r} = P_{w} - \boldsymbol{t}, \\
		\Phi_{RP\to CP}: \, P_{c} = R^{-1} P_{r},           \\
		\Phi_{CP\to D}: \, D = \|P_{c}\|_{2}
	\end{cases}
	\label{eq:trans}
\end{equation}
Here $D$ is the depth map of panoramic image $I$, $R$ and $\boldsymbol{t}$ are the rotation matrix and translation vector of the camera pose, respectively.
In the forward direction, the camera point map $P_{c} = D \odot P_{u}$ scales the unit sphere by depth,
$P_{r} = R P_{c}$ rotates into the reference frame, and $P_{w} = P_{r} + \boldsymbol{t}$ translates to world coordinates; the inverse direction reverses these steps.
In $\Phi_{CP\to D}$, the angular reprojection error is neglected for efficiency, as its influence on results is negligible.
\subsection{Metric Prediction}
Unlike perspective cameras where focal length and depth are inherently coupled,
the fixed ERP projection model eliminates intrinsic ambiguity and thus simplifies the model's learning of metric depth.
Combined with the consistent metric scale across our dataset (from LiDAR capture and physically based rendering),
this enables direct metric prediction without a separate scale estimation module~\cite{keetha2025mapanything, wang2025moge2}.
We simply normalize the ground-truth scale (e.g., divide by $s=10$) during supervision,
which yields stable and accurate metric predictions.
\subsection{Loss Function}
\label{sec:loss}
We train Argus end-to-end using multiple losses, including a covisibility loss, camera loss, depth loss, multiple point map losses, and a geometry joint loss.
\par\noindent\textbf{Covisibility Loss.} The covisibility loss $\mathcal{L}_{\text{covis}}$ is formulated as:
\begin{equation}
	\mathcal{L}_{\text{covis}} = -\frac{1}{N} \sum_{i=1}^N \left[ \mathbf{C}_i \log(\sigma(\hat{\mathbf{C}}_i)) + (1-\mathbf{C}_i)\log(1-\sigma(\hat{\mathbf{C}}_i)) \right]
	\label{eq:loss_covis}
\end{equation}
where $N$ is the sequence length and $\sigma(\cdot)$ is the sigmoid function.
\par\noindent\textbf{Camera Loss.}
The camera loss employs two terms: the quaternion-based rotation loss $\mathcal{L}_{q}$ and the translation loss $\mathcal{L}_{t}$.
These measure the L1 distance between predicted and ground-truth quaternions $\hat{q}_i$, $q_i$,
and between scaled predicted and scaled ground-truth translation vectors $\hat{t}_i$, $\frac{t_i}{s}$.
Unlike previous methods~\cite{wang2025vggt, keetha2025mapanything},
additional confidence scores $\mathcal{C}^{q}_i$ and $\mathcal{C}^{t}_i$ are predicted respectively for rotation and translation to improve usability.
These scores reflect the model's pose estimation confidence and are integrated into the corresponding loss terms as follows:
\begin{equation}
	\mathcal{L}_{q} = \frac{1}{N} \sum_{i=1}^N (\mathcal{C}^{q}_i + 1) \cdot\lVert\hat{q}_i - q_i\rVert_1 -\alpha \log \mathcal{C}^{q}_i
	\label{eq:loss_q}
\end{equation}
\begin{equation}
	\mathcal{L}_{t} = \frac{1}{N} \sum_{i=1}^N (\mathcal{C}^{t}_i + 1) \cdot \lVert\hat{t}_i - \frac{t_i}{s}\rVert_1-\alpha \log \mathcal{C}^{t}_i
	\label{eq:loss_t}
\end{equation}
The overall camera loss is formulated as: $\mathcal{L}_{cam} = \mathcal{L}_{q} + \mathcal{L}_{t}$.
\par\noindent\textbf{Depth Loss.}
The depth loss follows~\cite{wang2025vggt},
adopting an aleatoric uncertainty loss that weights the L2 between the predicted depth $\hat D_{i,j}$ and the ground-truth depth $D_{i,j}$ with the predicted uncertainty map $C^{D}_{i,j}$.
The $\mathcal{L}_{d}$ is formulated as:
\begin{equation}
	\mathcal{L}_{d} = \frac{1}{NHW} \sum_{i=1}^N \sum_{j=1}^{HW}(C^{D}_{i,j} + 1)\left [ \left\lVert \hat D_{i,j}-\frac{D_{i,j}}{s}\right\rVert + \left\lVert \nabla \hat D_{i,j}-\nabla \frac{D_{i,j}}{s}\right\rVert \right ]  - \alpha \log C^{D}_{i,j}
	\label{eq:loss_d}
\end{equation}
where $\nabla$ is the gradient operator.
\par\noindent\textbf{Point Map Loss.}
The point map loss $\mathcal{L}_{p}$ is analogous to the depth loss.
\begin{equation}
	\begin{split}
		\mathcal{L}_{p} = \frac{1}{3NHW} \sum_{i=1}^N \sum_{j=1}^{HW} (C^{P}_{i,j} + 1) \Big[ \left\lVert \hat P_{i,j}-\frac{P_{i,j}}{s}\right\rVert \\
			+ \left\lVert \mathbf{n}(\hat P_{i,j})-\mathbf{n}\!\left(\frac{P_{i,j}}{s}\right) \right\rVert \Big] - \alpha \log C^{P}_{i,j}
	\end{split}
	\label{eq:loss_p}
\end{equation}
Note that $\mathbf{n}(\cdot)$ denotes the operator computing point normals.
Substituting predicted point maps under distinct spatial coordinate systems into \cref{eq:loss_p} yields
$\mathcal{L}_{cp} = \mathcal{L}_{p}(P_c)$, $\mathcal{L}_{rp} = \mathcal{L}_{p}(P_r)$ and $\mathcal{L}_{wp} = \mathcal{L}_{p}(P_w)$, respectively.
\par\noindent\textbf{Geometry Joint Loss.}
To promote mutual enhancement across geometric prediction branches,
we further supervise adjacent transformations in all forward and inverse steps of pixel-to-world coordinate mapping.
This is formulated as:
\begin{equation}
	\begin{aligned}
		\mathcal{L}_{\text{joint}} & = \mathcal{L}_{\text{p}}\bigl[ \Phi_{D\to CP}( \hat D ) \bigr] + \mathcal{L}_{\text{p}}\bigl[ \Phi_{CP\to RP}( \hat P_c, \hat R ) \bigr] + \mathcal{L}_{\text{p}}\bigl[ \Phi_{RP\to WP}( \hat P_r, \hat t ) \bigr]         \\
		                           & \quad + \mathcal{L}_{\text{p}}\bigl[ \Phi_{WP\to RP}( \hat P_w, \hat t ) \bigr] + \mathcal{L}_{\text{p}}\bigl[ \Phi_{RP\to CP}( \hat P_r, \hat R ) \bigr] + \mathcal{L}_{\text{d}}\bigl[ \Phi_{CP\to D}( \hat P_c ) \bigr]
	\end{aligned}
	\label{eq:loss_joint}
\end{equation}%
Each term takes the prediction from one branch, applies the predicted geometric transform to derive the next-stage representation, and supervises the result against the corresponding ground-truth, thereby enforcing cross-branch geometric consistency.
Note that all inputs to the joint loss ($\hat D$, $\hat P_c$, $\hat P_r$, $\hat P_w$, $\hat{R}$, $\hat{\boldsymbol{t}}$) are network predictions from their respective branches, ensuring gradient flow across all branches for mutual facilitation.
\par\noindent\textbf{Total Loss.}
The final weighted loss $\mathcal{L}$ is as follows:
\begin{equation}
	\mathcal{L}= 0.1 \cdot \mathcal{L}_{covis} + 5.0 \cdot \mathcal{L}_{cam} + \mathcal{L}_{d} + \mathcal{L}_{cp} + \mathcal{L}_{rp} + \mathcal{L}_{wp} + \mathcal{L}_{joint}
	\label{eq:loss_total}
\end{equation}
\section{Implementation Details}
We train Argus by optimizing the training loss with the AdamW~\cite{loshchilov2017decoupled} optimizer for 99K iterations.
We use a cosine learning rate scheduler with a peak learning rate of $5e-5$ and a warmup of 9.9K iterations.
For every batch, we randomly sample 2 to 28 views from a random training scene.
Since the covisibility adjacency matrix between views in each scene is precomputed,
we ensure the sampled views are connected during training.
Besides, we randomly split both real and synthetic subsets into training and test sets at a 9:1 scene-level ratio.
Due to the non-uniform pixel distribution of ERP panoramic images,
pixel utilization at the north and south poles is extremely low.
We first resize the panoramic images, depths and point maps to 560$\times$280,
then crop the top and bottom 15\% of pixels each,
yielding a final resolution of 560$\times$196.
We also randomly apply color jittering, Gaussian blur, and grayscale augmentation to the views.
In addition, a random rotation around the Y-axis is applied to the input of each panoramic viewpoint,
which corresponds to a horizontal pixel shift in the ERP image and augments viewpoint yaw diversity.
We set $\alpha=0.2$ as the weight of the regularization term for all confidence losses.
The training runs on 24 H20 GPUs with 141 GB memory over 36 hours.
We employ gradient norm clipping with a threshold of 1.0 to ensure training stability.
We leverage bfloat16 precision and gradient checkpointing to improve GPU memory and computational efficiency.
All evaluations are completed on a single H20 GPU.

\section{Benchmarking \& Results}
\label{sec:results}
\subsection{Baselines}
We benchmark Argus on a comprehensive suite of 3D vision geometry tasks using our Realsee3D dataset.
Given the long-standing scarcity of large-scale 3D panoramic datasets,
no off-the-shelf feed-forward network is available for direct panoramic 3D reconstruction.
For a fair comparison with SOTA methods,
we therefore finetune VGGT~\cite{wang2025vggt} (without tracking supervision~\cite{karaev2024cotracker,karaev2025cotracker3}),
MapAnything~\cite{keetha2025mapanything} (V1.1.1, using only image inputs) and $\pi^3$~\cite{wang2025pi} as our main baselines on Realsee3D.
All models share consistent experimental settings,
including the same train/test split,
data preprocessing,
and the same training iterations.
We denote these adapted models as VGGT360, MapAnything360 and $\pi^3$360.
Argus and VGGT360 are initialized from VGGT~\cite{wang2025vggt} pre-trained weights,
others from their own checkpoints.
It should be noted that the released $\pi^3$ pre-training weights are initialized from VGGT.
Realsee3D contains scenes represented by unordered images of arbitrary quantity.
For comprehensive scene evaluation,
validation uses all per-scene panoramic images,
differing from prior work~\cite{wang2025vggt} sampling fixed frames from ordered video streams.
Moreover, due to the challenging dataset,
traditional open-source SfM methods (COLMAP~\cite{schonberger2016structure} and OpenMVG~\cite{moulon2016openmvg}),
when applied out of the box,
are too slow and yield unusable results,
so they are omitted from comparison.
\subsection{Camera Pose Estimation}
We evaluate camera pose prediction on both subsets of Realsee3D,
reporting AUC, Relative Rotation Accuracy (RRA), and Relative Translation Accuracy (RTA) at thresholds of $5^{\circ}$, $10^{\circ}$, $20^{\circ}$,
Absolute Trajectory Error RMSE (ATE) for global pose accuracy,
and Acceptance Rate (A.R.)---the proportion of scenes where all poses have rotation error $<10^\circ$ and translation error $<0.5$\,m---to assess practical usability.
As shown in \cref{tab:pose},
Argus achieves the best metric pose accuracy (ATE and A.R.) on both subsets and leads in AUC on the synthetic subset.
Although $\pi^3$360 shows marginally higher relative metrics on the real subset---likely due to its stronger pre-training prior---Argus leads decisively in global metric accuracy and dominates all metrics on the synthetic subset where the domain gap from pre-training data is smaller.
Compared with MapAnything360 (which supports metric prediction), Argus reduces ATE by 28\% on the real subset and 69\% on the synthetic subset.
\begin{table}[tb]
	\tiny
	\caption{\textbf{Camera Pose Estimation on the Realsee3D Dataset.}}
	\label{tab:pose}
	\centering
	\begin{tabular}{@{}l*{12}{c}@{}}
		\toprule
		\multirow{2}*{Dataset Part} & \multirow{2}*{Method}                      & \multicolumn{3}{c}{AUC$\uparrow$} & \multicolumn{3}{c}{RRA$\uparrow$} & \multicolumn{3}{c}{RTA$\uparrow$} & \multirow{2}*{ATE$\downarrow$} & \multirow{2}*{A.R.$\uparrow$}                                                                                                       \\
		\cmidrule(lr){3-5}
		\cmidrule(lr){6-8}
		\cmidrule(lr){9-11}

		                            &                                            & @$5^{\circ}$                      & @$10^{\circ}$                     & @$20^{\circ}$                     & @$5^{\circ}$                   & @$10^{\circ}$                 & @$20^{\circ}$   & @$5^{\circ}$   & @$10^{\circ}$  & @$20^{\circ}$                                   \\
		\midrule
		\multirow{4}*{Real}
		                            & VGGT360\cite{wang2025vggt}                 & 67.89                             & 83.00                             & 91.00                             & 96.88                          & 99.10                         & 99.46           & 97.37          & 99.05          & 99.66          & --             & --            \\
		                            & MapAnything360\cite{keetha2025mapanything} & \textbf{72.66}                    & 85.37                             & 92.45                             & \textbf{99.41}                 & 99.94                         & \textbf{100.00} & 95.86          & 98.97          & 99.77          & 0.134          & 94.0          \\
		                            & $\pi^3$360\cite{wang2025pi}                & 72.39                             & \textbf{85.92}                    & \textbf{92.90}                    & 98.60                          & \textbf{100.00}               & \textbf{100.00} & \textbf{98.58} & \textbf{99.74} & \textbf{99.92} & --             & --            \\
		                            & Ours                                       & 71.88                             & 85.52                             & 92.64                             & 98.30                          & \textbf{100.00}               & \textbf{100.00} & 98.10          & 99.55          & 99.83          & \textbf{0.096} & \textbf{98.2} \\
		\midrule
		\multirow{4}*{Synthetic}
		                            & VGGT360\cite{wang2025vggt}                 & 91.10                             & 95.38                             & 97.60                             & 99.67                          & 99.86                         & 99.92           & 99.57          & 99.80          & 99.89          & --             & --            \\
		                            & MapAnything360\cite{keetha2025mapanything} & 90.84                             & 95.11                             & 97.42                             & 99.46                          & 99.78                         & 99.92           & 99.20          & 99.68          & 99.83          & 0.087          & 99.3          \\
		                            & $\pi^3$360\cite{wang2025pi}                & 93.46                             & 96.58                             & 98.20                             & 99.72                          & 99.88                         & 99.91           & 99.62          & \textbf{99.82} & 99.89          & --             & --            \\
		                            & Ours                                       & \textbf{94.44}                    & \textbf{97.29}                    & \textbf{98.63}                    & \textbf{99.76}                 & \textbf{99.90}                & \textbf{99.95}  & \textbf{99.65} & \textbf{99.82} & \textbf{99.90} & \textbf{0.027} & \textbf{99.8} \\

		\bottomrule
	\end{tabular}
\end{table}
\begin{table}[tb]
	\tiny
	\caption{\textbf{Multi-view Depth Estimation on the Realsee3D Dataset.}}
	\label{tab:mvdepth}
	\centering
	\begin{tabular}{@{}l*{11}{c}@{}}
		\toprule
		\multirow{2}*{Alignment} & \multirow{2}{*}{Method}                    & \multicolumn{5}{c}{Real} & \multicolumn{5}{c}{Synthetic}                                                                                                                                                                           \\
		\cmidrule(lr){3-7}
		\cmidrule(lr){8-12}
		                         &                                            & AbsRel$\downarrow$       & $\delta_{1}$$\uparrow$        & $\delta_{2}$$\uparrow$ & RMSE$\downarrow$ & MAE$\downarrow$ & AbsRel$\downarrow$ & $\delta_{1}$$\uparrow$ & $\delta_{2}$$\uparrow$ & RMSE$\downarrow$ & MAE$\downarrow$ \\
				\midrule
				\multirow{4}*{IRLS}
		                         & VGGT360\cite{wang2025vggt}                 & 0.051                    & 64.22                         & 96.21                  & 0.445            & 0.119           & 0.025              & 86.60                  & 98.78                  & 0.158            & 0.035           \\
		                         & MapAnything360\cite{keetha2025mapanything} & 0.063                    & 54.75                         & 95.72                  & 0.541            & 0.126           & 0.034              & 76.47                  & 98.65                  & 0.165            & 0.042           \\
		                         & $\pi^3$360\cite{wang2025pi}                & 0.053                    & 58.63                         & 95.87                  & 0.454            & 0.129           & 0.033              & 78.14                  & 98.25                  & 0.192            & 0.052           \\
		                         & Ours                                       & \textbf{0.048}           & \textbf{70.22}                & \textbf{96.51}         & \textbf{0.401}   & \textbf{0.102}  & \textbf{0.019}     & \textbf{91.42}         & \textbf{99.03}         & \textbf{0.138}   & \textbf{0.027}  \\
				\midrule
				\multirow{4}*{Median}
		                         & VGGT360\cite{wang2025vggt}                 & 0.056                    & 57.69                         & 96.08                  & 0.447            & 0.127           & 0.029              & 85.27                  & 98.76                  & 0.159            & 0.038           \\
		                         & MapAnything360\cite{keetha2025mapanything} & 0.066                    & 51.08                         & 95.90                  & 0.563            & 0.132           & 0.037              & 74.62                  & 98.66                  & 0.186            & 0.046           \\
		                         & $\pi^3$360\cite{wang2025pi}                & 0.060                    & 51.85                         & 95.65                  & 0.456            & 0.140           & 0.034              & 75.96                  & 98.22                  & 0.194            & 0.055           \\
		                         & Ours                                       & \textbf{0.050}           & \textbf{65.81}                & \textbf{96.60}         & \textbf{0.406}   & \textbf{0.107}  & \textbf{0.019}     & \textbf{90.75}         & \textbf{99.02}         & \textbf{0.139}   & \textbf{0.029}  \\
		\midrule
		\multirow{2}*{Metric}
		                         & MapAnything360\cite{keetha2025mapanything} & 0.070                    & 46.31                         & 95.79                  & 0.567            & 0.137           & 0.063              & 74.64                  & 98.64                  & 0.185            & 0.044           \\
		                         & Ours                                       & \textbf{0.050}           & \textbf{66.28}                & \textbf{96.62}         & \textbf{0.407}   & \textbf{0.106}  & \textbf{0.035}     & \textbf{91.24}         & \textbf{99.02}         & \textbf{0.139}   & \textbf{0.027}  \\

		\bottomrule
	\end{tabular}
\end{table}
\subsection{Depth Estimation}
We benchmark multi-view depth estimation on the Realsee3D dataset using
Absolute Relative Error (AbsRel), Root Mean Squared Error (RMSE), Mean Absolute Error (MAE),
and interior percentage metrics $\delta_{<1.03}$ ($\delta_{1}$) and $\delta_{<1.25}$ ($\delta_{2}$).
As several baselines cannot predict metric depth, ground-truth alignment is necessary before evaluation.
For fair comparison with all methods, we report results under three alignment schemes:
Iterative Reweighted Least Squares (IRLS) alignment~\cite{kummerle2021iteratively},
median alignment,
and absolute metric evaluation without any alignment.
Results are given in \cref{tab:mvdepth}.
Argus outperforms all baselines in all metrics.
We further evaluate Argus on monocular depth estimation for Realsee3D,
and test its zero-shot generalization on two standard benchmarks: Matterport3D~\cite{chang2017matterport3d} and Stanford2D3D~\cite{armeni2017joint}.
Although not specifically trained for monocular depth estimation,
our method still achieves highly competitive performance.
Please refer to supplementary materials for the detailed results.
\subsection{Point Map Reconstruction}
\begin{table}[tb]
	\tiny
	\caption{\textbf{Point Map Reconstruction on the Realsee3D Dataset.}}
	\label{tab:point}
	\centering
	\begin{tabular}{@{}l*{13}{c}@{}}
		\toprule
		\multirow{3}*{Alignment} & \multirow{3}*{Method}                      & \multicolumn{6}{c}{Real}             & \multicolumn{6}{c}{Synthetic}                                                                                                                                                                                                                                                                        \\
		\cmidrule(lr){3-8}
		\cmidrule(lr){9-14}
		                         &                                            & \multicolumn{2}{c}{Acc.$\downarrow$} & \multicolumn{2}{c}{Comp.$\downarrow$} & \multicolumn{2}{c}{N.C.$\uparrow$} & \multicolumn{2}{c}{Acc.$\downarrow$} & \multicolumn{2}{c}{Comp.$\downarrow$} & \multicolumn{2}{c}{N.C.$\uparrow$}                                                                                                       \\
		\cmidrule(lr){3-4}
		\cmidrule(lr){5-6}
		\cmidrule(lr){7-8}
		\cmidrule(lr){9-10}
		\cmidrule(lr){11-12}
		\cmidrule(lr){13-14}
		                         &                                            & Mean                                 & Med.                                  & Mean                               & Med.                                 & Mean                                  & Med.                               & Mean           & Med.           & Mean           & Med.           & Mean           & Med.           \\
		\midrule
		\multirow{4}*{ICP}
		                         & VGGT360\cite{wang2025vggt}                 & \textbf{0.058}                       & 0.039                                 & 0.074                              & \textbf{0.034}                       & 0.884                                 & 0.981                              & 0.023          & 0.014          & 0.016          & 0.011          & 0.923          & 0.995          \\
		                         & MapAnything360\cite{keetha2025mapanything} & 0.082                                & 0.054                                 & 0.127                              & 0.095                                & 0.829                                 & 0.937                              & 0.038          & 0.026          & 0.053          & 0.043          & 0.880          & 0.982          \\
		                         & $\pi^3$360\cite{wang2025pi}                & \textbf{0.058}                       & 0.040                                 & 0.078                              & \textbf{0.034}                       & 0.868                                 & 0.975                              & 0.030          & 0.017          & 0.018          & 0.013          & 0.888          & 0.991          \\
		                         & Ours                                       & \textbf{0.058}                       & \textbf{0.037}                        & \textbf{0.062}                     & \textbf{0.034}                       & \textbf{0.894}                        & \textbf{0.983}                     & \textbf{0.018} & \textbf{0.011} & \textbf{0.014} & \textbf{0.010} & \textbf{0.943} & \textbf{0.997} \\
		\midrule
		\multirow{2}*{Metric}
		                         & MapAnything360\cite{keetha2025mapanything} & 0.089                                & 0.057                                 & 0.075                              & 0.047                                & 0.825                                 & 0.962                              & 0.039          & 0.023          & 0.025          & 0.016          & 0.873          & 0.991          \\
		                         & Ours                                       & \textbf{0.056}                       & \textbf{0.037}                        & \textbf{0.063}                     & \textbf{0.034}                       & \textbf{0.894}                        & \textbf{0.986}                     & \textbf{0.020} & \textbf{0.012} & \textbf{0.015} & \textbf{0.010} & \textbf{0.939} & \textbf{0.997} \\

		\bottomrule
	\end{tabular}
\end{table}
\begin{figure}[tb]
	\centering
	\includegraphics[width=1\linewidth]{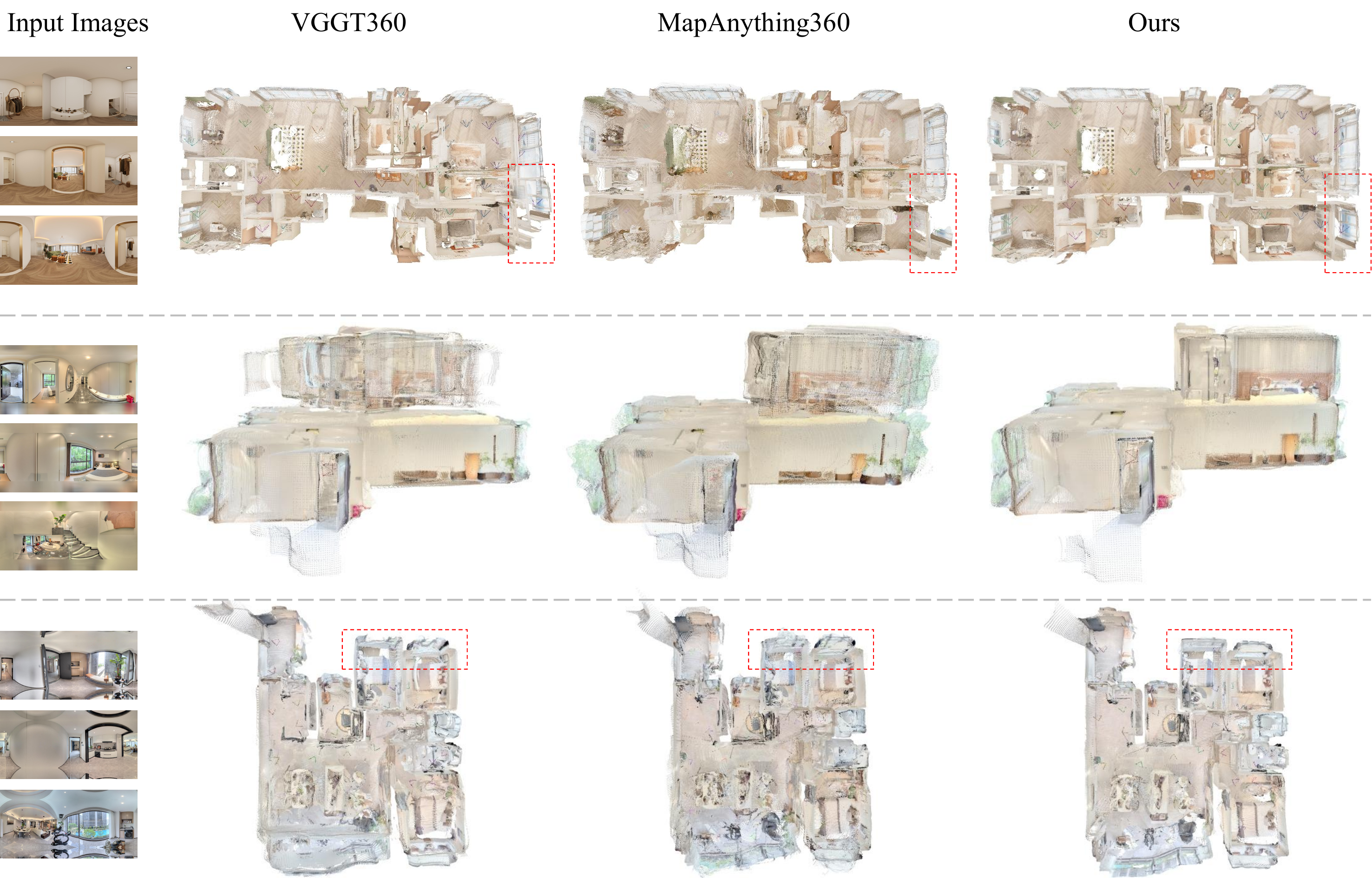}

	\caption{
		\textbf{Qualitative Comparison.}
	}
	\label{fig:comparison}
\end{figure}
We evaluate scene-level point map reconstruction in the world coordinate system on our Realsee3D dataset, reporting Accuracy (Acc.), Completeness (Comp.), and Normal Consistency (N.C.) in \cref{tab:point}.
We present results under two alignment schemes:
alignment using the Umeyama algorithm~\cite{umeyama1991least} followed by Iterative Closest Point (ICP) refinement~\cite{besl1992method},
and absolute metric evaluation without any alignment.
Argus achieves overall superior alignment performance and holds a clear absolute advantage under metric evaluation.
The qualitative comparison of reconstructions is shown in \cref{fig:comparison},
Argus yields more accurate metric geometry and sharper structural boundaries than previous methods.
\begin{table}[tb]
	\caption{\textbf{Runtime and Peak GPU Memory Usage of Argus with Varying Input Frame Counts.}}
	\label{tab:Runtime}
	\centering
	\begin{tabular}{@{}l*{8}{c}@{}}
		\toprule
		\# Image Number & 1    & 2    & 4    & 8    & 16   & 32   & 64   & 128   \\
		\midrule
		Runtime (s)     & 0.11 & 0.13 & 0.18 & 0.32 & 0.66 & 1.47 & 3.67 & 10.55 \\
		Memory (GB)     & 2.05 & 2.24 & 2.60 & 3.38 & 3.54 & 3.88 & 4.55 & 7.22  \\
		\bottomrule
	\end{tabular}
\end{table}
\subsection{Efficiency Evaluation}
The runtime and peak GPU memory usage across different numbers of input frames are reported in \cref{tab:Runtime}.
In practice, we run inference in BF16 precision and perform on-the-fly cleanup of redundant intermediate features,
which keeps the peak memory footprint at a favorable level even when the number of input images exceeds 100.
Note that the reported data ignore the memory of loading the pretrained model (4.64GB).
Our model contains 1.31 billion parameters.
Notably,
the multiple DPT heads for cross-coordinate point map prediction can also be disabled during inference,
as optimal performance is typically achieved with only the pose and depth heads.
Thus, our multi-head prediction and supervision scheme markedly boosts geometric learning during training,
with zero additional inference overhead.
\subsection{Visualization}
\par\noindent\textbf{Reference View Learning.}
\begin{figure}[tb]
	\centering
	\includegraphics[width=1\linewidth]{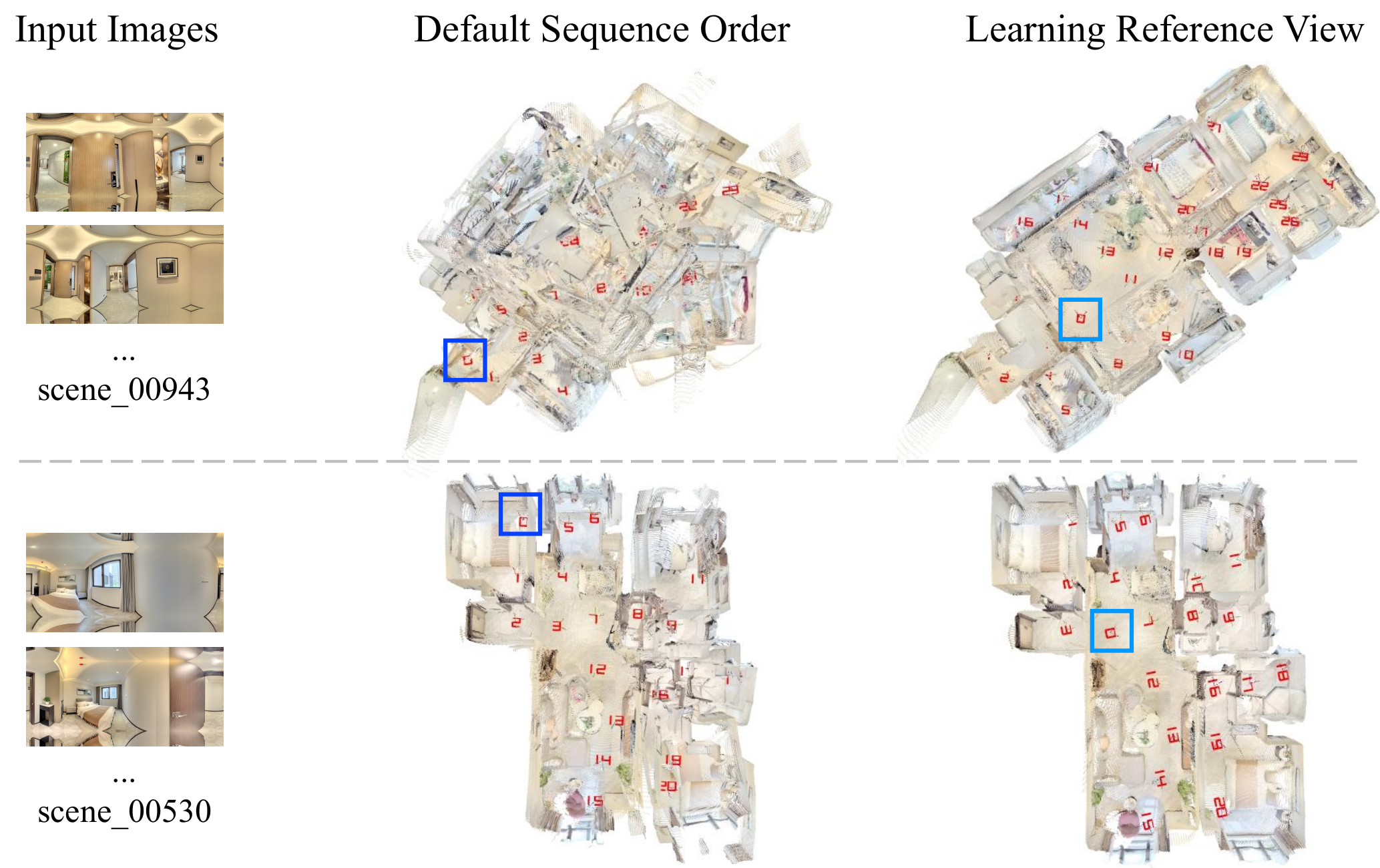}

	\caption{
		\textbf{Visualization of the Effectiveness of Reference View Learning.}
	}
	\label{fig:ref_vis}
\end{figure}
As shown in \cref{fig:ref_vis}, real-world capture often starts from arbitrary positions,
and the initial viewpoint is frequently located at scene corners or boundaries with low covisibility,
which greatly raises the difficulty of robust reconstruction.
With global covisibility supervision,
Argus selects a superior initial reference view,
typically near the scene center with stronger direct or indirect connections to other viewpoints.
This improves scene-level reconstruction robustness by reducing drift and accumulated errors.
More importantly, it enables metric prediction in the absolute world coordinate frame.
\par\noindent\textbf{Generalizability.}
\begin{figure}[tb]
	\centering
	\includegraphics[width=1\linewidth]{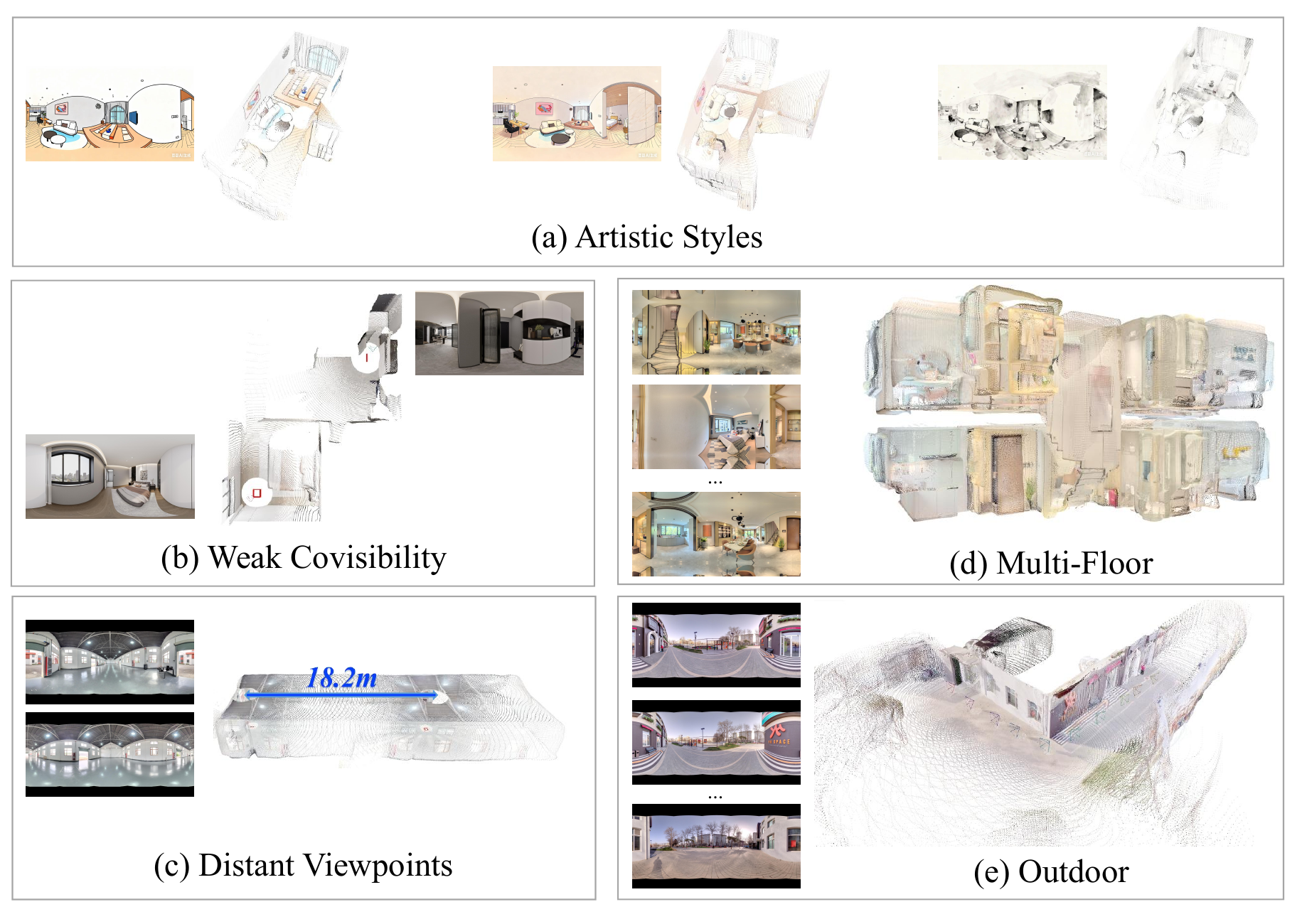}

	\caption{
		\textbf{Visual Samples with Generalization Ability.}
	}
	\label{fig:vis}
\end{figure}
As shown in \cref{fig:vis},
Argus generalizes strongly beyond the training distribution.
In particular,
it maintains reliable metric 3D reconstruction for input panoramas with diverse AI-generated artistic styles~\cite{seedream2025seedream} (e.g., sketches, cartoons).
Furthermore, Argus exhibits strong robustness under challenging capture settings:
weak covisibility, multi-floor indoor scenes, long-range viewpoints, and small-scale outdoor environments.
\subsection{Ablation Studies}
\begin{table}[tb]
	\scriptsize
	\caption{\textbf{Ablation Studies on the Synthetic Subset of Realsee3D Dataset.} }
	\label{tab:ablation_synthetic}
	\centering
	\begin{tabular}{@{}l*{10}{c}@{}}
		\toprule
		Method                                                             & AbsRel$\downarrow$ & $\delta_{1}$$\uparrow$ & $\delta_{2}$$\uparrow$ & RMSE$\downarrow$ & MAE$\downarrow$ & Acc.$\downarrow$ & Comp.$\downarrow$ & N.C.$\uparrow$ & ATE$\downarrow$ \\
				\midrule
		Full                                                               & \textbf{0.035}     & \textbf{91.24}         & \textbf{99.02}         & \textbf{0.139}   & \textbf{0.027}  & \textbf{0.020}   & \textbf{0.015}    & \textbf{0.939} & \textbf{0.027}  \\
		w/o reference view learning                                        & 0.037              & 90.45                  & 98.96                  & 0.142            & 0.029           & 0.024            & 0.017             & 0.923          & 0.060           \\
		w/o $\mathcal{L}_{joint}$                                          & 0.047              & 90.71                  & 98.86                  & 0.158            & 0.030           & \textbf{0.020}   & \textbf{0.015}    & 0.937          & 0.049           \\
		w/o $\mathcal{L}_{cp}$                                             & 0.055              & 89.90                  & 98.79                  & 0.161            & 0.031           & \textbf{0.020}   & \textbf{0.015}    & 0.936          & 0.047           \\
		w/o $\mathcal{L}_{rp}$                                             & 0.045              & 90.55                  & 98.83                  & 0.159            & 0.030           & \textbf{0.020}   & \textbf{0.015}    & 0.937          & 0.051           \\
		w/o $\mathcal{L}_{cp}$ \& $\mathcal{L}_{rp}$                       & 0.055              & 89.75                  & 98.79                  & 0.160            & 0.031           & \textbf{0.020}   & \textbf{0.015}    & 0.936          & 0.048           \\
		w/o $\mathcal{L}_{cp}$ \& $\mathcal{L}_{rp}$ \& $\mathcal{L}_{wp}$ & 0.075              & 88.52                  & 98.60                  & 0.169            & 0.034           & --               & --                & --             & 0.061           \\
		\bottomrule
	\end{tabular}
\end{table}
Ablation studies are presented in~\cref{tab:ablation_synthetic}.
We report metric prediction results on the synthetic subset with key designs progressively ablated.
Removing reference view learning more than doubles the ATE (0.027$\to$0.060),
confirming its critical role in establishing a consistent global coordinate frame.
The geometric joint loss and multi-coordinate point map supervision improve both depth and pose accuracy:
removing $\mathcal{L}_{joint}$ raises AbsRel from 0.035 to 0.047 and ATE from 0.027 to 0.049,
and progressively dropping point map losses further degrades performance,
with the removal of all three causing the largest decline (AbsRel 0.075, RMSE 0.169).
Notably, removing $\mathcal{L}_{cp}$ alone has a larger impact than removing $\mathcal{L}_{rp}$ alone,
suggesting that camera-coordinate supervision provides a stronger learning signal for depth.
Although $P_c$ and $D$ are deterministically interconvertible,
they represent geometry in different dimensional spaces and induce significantly different gradient dynamics during optimization.
\section{Conclusions, Limitations, and Future Work}
\label{sec:conclusions}
We present Argus, a feed-forward model that achieves metric panoramic 3D reconstruction from unordered indoor images in a single forward pass,
together with Realsee3D,
a hybrid indoor benchmark of 10K real and synthetic scenes with 299K panoramic viewpoints and precise metric annotations.
Argus introduces covisibility-guided reference view selection to robustly anchor unordered inputs,
and a panoramic geometric factorization scheme that decomposes pixel-to-world mapping into independently supervised intermediate representations with cross-coordinate joint constraints.
These designs enable Argus to achieve state-of-the-art overall performance among feed-forward methods on camera pose estimation,
metric depth prediction, and point map reconstruction, while maintaining favorable efficiency.
Argus excels in structured indoor environments but has clear limitations:
its indoor-focused training data constrains zero-shot generalization to unbounded outdoor and aerial scenes,
and memory scaling limits reconstruction from very large panorama sets (hundreds to thousands of views),
making scale-agnostic reconstruction a promising future direction.
More discussions are provided in the supplementary material.
We believe Realsee3D and Argus together provide the community with a strong foundation for advancing metric panoramic 3D reconstruction and its downstream applications.

%
%
\bibliographystyle{splncs04}
\bibliography{main}

\clearpage
\setcounter{page}{1}
\setcounter{section}{0}
\begin{center}
	{\Large\bfseries Supplementary Material}
\end{center}
\renewcommand{\thesection}{\Alph{section}}
\begin{figure}[H]
	\centering
	\includegraphics[width=1\linewidth]{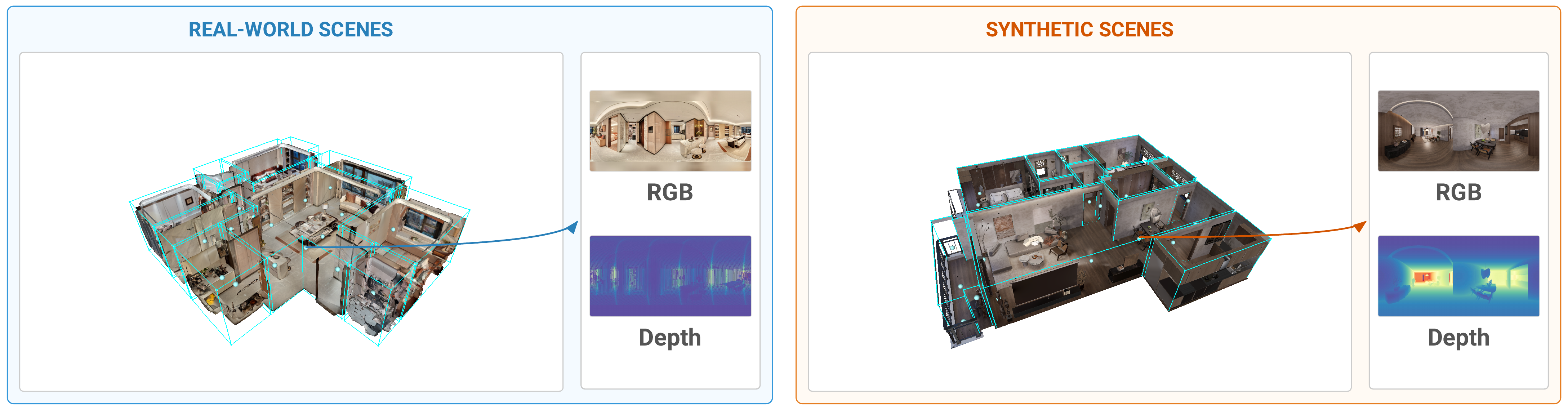}
	\caption{
		\textbf{Examples of Realsee3D.}
	}
	\label{fig:realsee3d_overview}
\end{figure}

\section{More Dataset Details}
\subsection{Dataset Characteristics}
Realsee3D is distinguished by several key features:
\begin{itemize}
	\item \textbf{Panoramic Format \& Validity Masks:} Visual data is provided as high-resolution $1600 \times 800$ ERP panoramic images. While synthetic scenes offer a complete $360^{\circ} \times 180^{\circ}$ field of view, real-world captures inherently contain blind spots at the zenith and nadir due to hardware constraints. We provide explicit binary validity masks to account for these unobserved regions.
	\item \textbf{Sparse vs. Dense Depth Profiles:} Reflecting its hybrid nature, Realsee3D offers two distinct depth profiles.
	      As shown in \cref{fig:realsee3d_overview}, real-world data provides sparse metric depth inherently constrained by LiDAR discrete sampling, while synthetic scenes offer dense, continuous depth maps. This duality allows researchers to evaluate algorithm robustness across varying depth qualities. Both modalities include 16-bit relative depth maps and scalar factors to recover absolute metric depth.
	\item \textbf{Spatial and Geometric Metadata:} Each viewpoint includes a $4 \times 4$ Camera-to-World transformation matrix defining its 6-DoF absolute pose. For complex, multi-story environments, we provide explicit floor-level indices to facilitate vertical semantic understanding. Additionally, synthetic scenes include pairwise covisibility scores, indicating the visual overlapping area between views.
	\item \textbf{Unordered Sequences:} Reflecting real-world usage where users may capture views in an arbitrary order, the panoramic sequences in Realsee3D are unordered. This presents a unique challenge for reconstruction algorithms, which must robustly handle varying overlap and connectivity.
\end{itemize}
\subsection{Dataset Diversity and Statistics}
In this section, we provide a detailed statistical analysis of the Realsee3D dataset to highlight its diversity across both real-world and synthetic subsets.
\par\noindent\textbf{Viewpoint Distance.}
In the Realsee3D dataset, the average pairwise distances between viewpoints are 5.20 meters for the real subset and 5.93 meters for the synthetic subset.
\par\noindent\textbf{Depth Distribution.}
\begin{figure}[tb]
	\centering
	\includegraphics[width=1\linewidth]{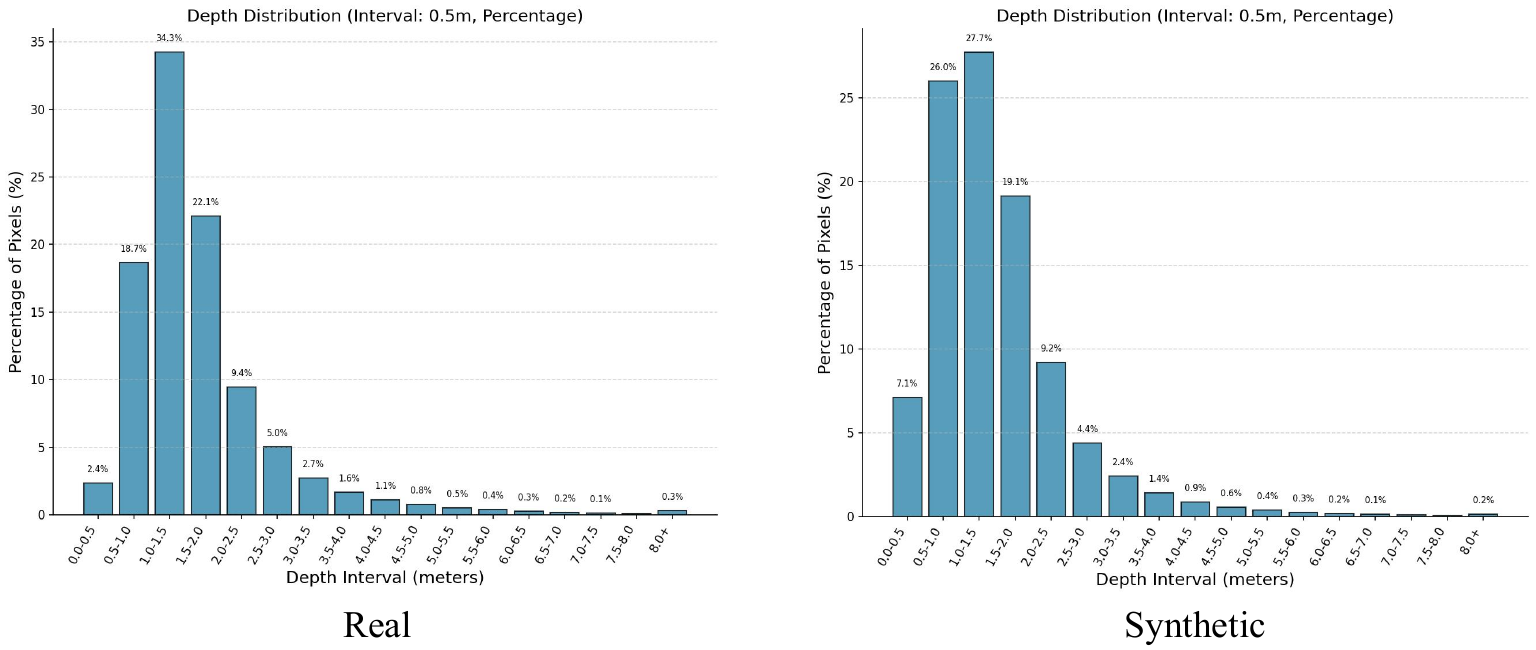}
	\caption{
		\textbf{Depth Distribution of Realsee3D.}
	}
	\label{fig:depth_dist}
\end{figure}
The depth distribution of the two subsets of the Realsee3D dataset is shown in \cref{fig:depth_dist}.
The average absolute depths of the real subset and the synthetic subset are 1.679 and 1.495 meters, respectively.
\par\noindent\textbf{Room Type Distribution.}
Realsee3D covers a wide range of residential room types, reflecting the complexity and variety of real-world indoor environments.
As shown in \cref{tab:room_distribution},
the distribution of room types is highly consistent between the real and synthetic subsets,
with Bedroom, Bathroom, and Balcony being the most frequent categories.
Beyond these common areas,
the dataset also includes niche spaces such as Study Rooms, Cloakrooms, and functional areas like Nanny's Rooms,
ensuring broad coverage of diverse architectural layouts.
\begin{table}[tb]
	\centering
	\caption{\textbf{Room Type Distribution in Realsee3D.}
		The distribution is calculated over 9,483 rooms in the Real subset and 86,479 rooms in the Synthetic subset.}
	\label{tab:room_distribution}
	\footnotesize
	\begin{tabular}{lrr|lrr}
		\toprule
		Room Type   & Real (\%) & Synth (\%) & Room Type     & Real (\%) & Synth (\%) \\
		\midrule
		Bedroom     & 32.59     & 33.07      & Dining Room   & 1.00      & 1.98       \\
		Bathroom    & 21.61     & 18.99      & Cloakroom     & 0.98      & 1.12       \\
		Balcony     & 10.70     & 14.38      & Study Room    & 1.64      & 0.95       \\
		Living Room & 10.46     & 10.40      & Entrance Hall & 1.20      & 0.89       \\
		Kitchen     & 9.57      & 10.09      & Storage Room  & 0.89      & 2.04       \\
		Corridor    & 6.94      & 5.58       & Other         & 0.42      & 0.51       \\
		\bottomrule
	\end{tabular}
\end{table}
\begin{table}[tb]
	\centering
	\caption{\textbf{Distribution of Architectural Elements.}
		The counts represent the total number of occurrences across the entire dataset (10k scenes).}
	\label{tab:element_distribution}
	\footnotesize
	\begin{tabular}{lrr}
		\toprule
		Element Type      & Real (Count) & Synthetic (Count) \\
		\midrule
		Single Door       & 8,924        & 106,655           \\
		Doorway/Opening   & 6,756        & 25,591            \\
		Standard Window   & 4,331        & 50,217            \\
		Sliding Door      & 3,015        & 24,050            \\
		French Window     & 2,676        & 21,759            \\
		Single Bay Window & 1,655        & 10,301            \\
		Folding Door      & -            & 7,750             \\
		Double Door       & 72           & 350               \\
		\bottomrule
	\end{tabular}
\end{table}
\par\noindent\textbf{Architectural Elements.}
To evaluate geometric reconstruction and scene understanding,
Realsee3D incorporates a rich variety of architectural elements.
As summarized in \cref{tab:element_distribution},
the dataset contains a massive number of doors and windows with different styles and functionalities.
This includes standard single doors, sliding doors, French windows, and various types of bay windows.
The presence of these elements across 10,000 scenes provides challenging cases for identifying boundaries and understanding spatial connectivity.
\par\noindent\textbf{Scene Complexity.}
The average number of rooms per scene is 9.48 for the Real subset and 9.61 for the Synthetic subset,
demonstrating the large scale and complexity of the captured environments.
On average, each scene contains 24.2 (Real) to 30.5 (Synthetic) unique viewpoints,
providing dense multi-view coverage essential for robust 3D reconstruction.
Furthermore, while all synthetic scenes are single-story,
the Real subset includes 8.3\% multi-story environments (up to 3 floors),
adding another layer of vertical spatial complexity.
\section{Reference View Label Production}
\begin{figure}[tb]
	\centering
	\includegraphics[width=1\linewidth]{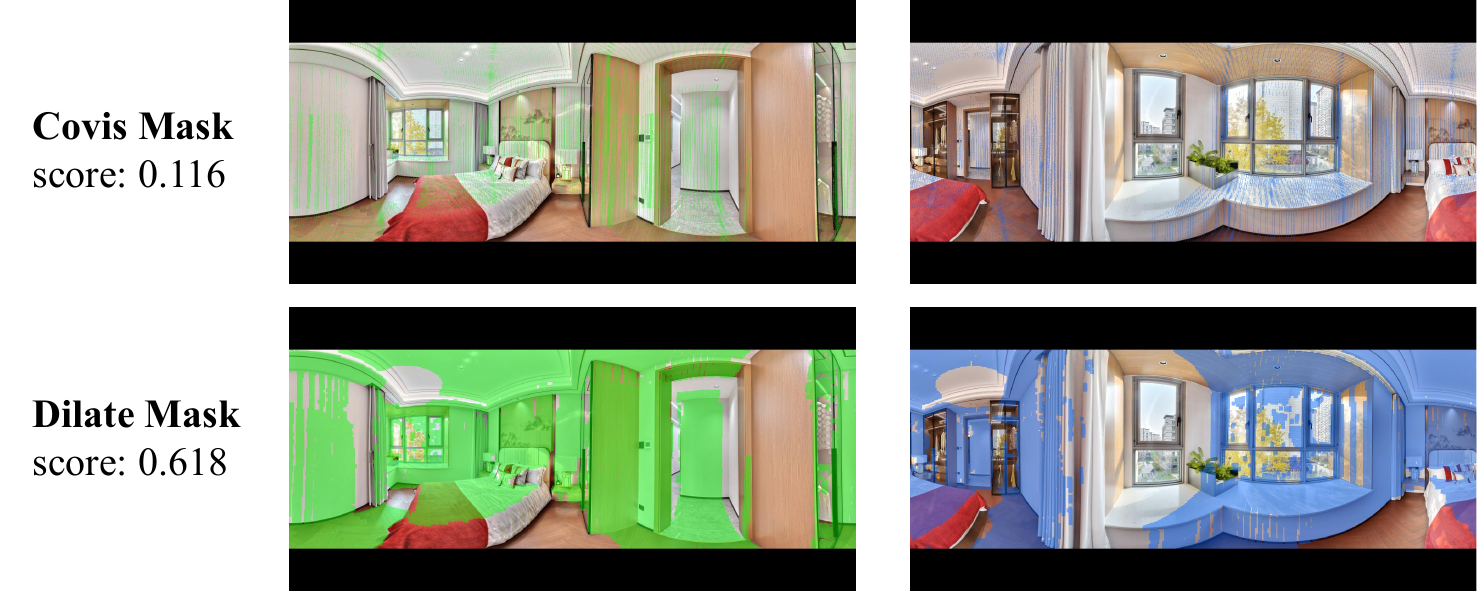}
	\caption{
		\textbf{Visualization of Masks Computed from Depth and Pose for Covisibility Scores.}
		We also dilate LiDAR-warped masks to alleviate sparsity and align real-synthetic score distribution.
	}
	\label{fig:make_covis}
\end{figure}

To enable the network to learn and select better reference views,
we generate supervised ground-truth through the following steps.
For any image pair $\left \langle I_i, I_j\right \rangle$ ($i,j \in \{1,2,\dots,N\}$),
$c_{ij}$ represents the co-visible pixel count calculated from relative pose and depth,
and $\mathcal{P}$ represents the total pixel count of a single image.
The covisibility scores are computed via reprojection with depth-buffer checking,
the visualization of the covisibility mask is illustrated in \cref{fig:make_covis}.
The covisibility score is defined as: $s_{ij} = \frac{c_{ij}}{\mathcal{P}}$.
Then we construct a dissimilarity matrix $\mathbf{M} \in \mathbb{R}^{N \times N}$,
where the higher the degree of common visibility between an image pair,
the closer their distance is deemed to be.
Thus, the covisibility score $s_{ij}$ is converted to a distance $m_{ij}$ with zero diagonal (a view has zero connectivity cost to itself) as:
\begin{equation}
	m_{ij} =
	\begin{cases}
		\frac{1}{s_{ij} + \epsilon} & , i \neq j \\
		0                           & , i = j
	\end{cases}
	\label{eq:d_ij}
\end{equation}
where $\epsilon$ is a small constant to avoid division by zero.
The dissimilarity matrix $\mathbf{M}$ is:
\begin{equation}
	\mathbf{M} = \begin{bmatrix}
		0      & m_{12} & \cdots & m_{1N} \\
		m_{21} & 0      & \cdots & m_{2N} \\
		\vdots & \vdots & \ddots & \vdots \\
		m_{N1} & m_{N2} & \cdots & 0
	\end{bmatrix}
	\label{eq:M_matrix}
\end{equation}
To quantify the minimum connectivity cost between all pairs of views,
we apply Dijkstra's algorithm~\cite{dijkstra1959note} to the dissimilarity matrix $\mathbf{M}$ to compute a shortest path matrix $\mathbf{P} \in \mathbb{R}^{N \times N}$.
We employ Dijkstra's algorithm instead of the standard Floyd-Warshall algorithm,
since our graph is sparse and Dijkstra's method yields higher computational efficiency.
We initialize $\mathbf{P}^{(0)} = \mathbf{M}$,
where $\mathbf{P}^{(t)}$ denotes the shortest path matrix at iteration $t$ (up to $N$ iterations, the total number of frames),
and $p_{ij}^{(t)}$ is the tentative shortest path length from view $i$ to view $j$ at iteration $t$.
We also define a binary visited mask $\mathbf{V} \in \left \{ 0,1\right \}^{N \times N}$ initialized to all zeros,
where $v_{ij}=1$ indicates that the shortest path from view $i$ to view $j$ has been determined.
For each iteration $t$,
we first identify the unvisited view $u_i$ with the minimum tentative distance from each source view $i$:
\begin{equation}
	u_i = \mathop{\arg\min}_{j \in \{1,2,\dots,N\}, \, v_{ij}=0} p_{ij}^{(t)}
	\label{eq:u_i}
\end{equation}
We then mark $u_i$ as visited for source view $i$ ($v_{iu_i}=1$)
and update the tentative shortest paths via the relaxation:
for all target views $j$,
the new path length pass $u_i$ is compared with the current tentative length, retaining the smaller value.
\begin{equation}
	p_{ij}^{(t+1)} = \min\left(p_{ij}^{(t)}, p_{iu_i}^{(t)} + m_{u_i j}\right), \quad \forall j \in \{1,2,\dots,N\}
	\label{eq:p_ij}
\end{equation}
The iteration terminates early if all nodes are visited (i.e., $\mathbf{V} = \mathbf{1}$) or if no reachable unvisited nodes remain
(all tentative distances are infinite).
The final shortest path matrix is defined as: $\mathbf{P} = \mathbf{P}^{(T)}$, where $T$ is the number of iterations until termination.
The element $p_{ij}$ in $\mathbf{P}$ thus represents the minimum distance (connectivity cost)
from view $i$ to view $j$ across all possible paths in the covisibility graph.
Finally, we select the reference view index $\mathcal{I}$ with the smallest total shortest path distance (highest global covisibility connectivity),
denoted as:
\begin{equation}
	\mathcal{I} = \mathop{\arg\min}_{i \in \{1,\ldots,N\}} \sum_{j=1}^N p_{ij}
	\label{eq:ref_index}
\end{equation}
\section{More Experiments}
\begin{table}[tb]
	\tiny
	\caption{\textbf{Monocular Depth Estimation Results on Realsee3D.}}
	\label{tab:mono_depth}
	\centering
	\begin{tabular}{@{}l*{11}{c}@{}}
		\toprule
		\multirow{2}*{Alignment} & \multirow{2}{*}{Method}                    & \multicolumn{5}{c}{Real} & \multicolumn{5}{c}{Synthetic}                                                                                                                                                                           \\
		\cmidrule(lr){3-7}
		\cmidrule(lr){8-12}
		                         &                                            & AbsRel$\downarrow$       & $\delta_{1}$$\uparrow$        & $\delta_{2}$$\uparrow$ & RMSE$\downarrow$ & MAE$\downarrow$ & AbsRel$\downarrow$ & $\delta_{1}$$\uparrow$ & $\delta_{2}$$\uparrow$ & RMSE$\downarrow$ & MAE$\downarrow$ \\
				\midrule
				\multirow{10}*{IRLS}
		                         & PanDA\cite{cao2025panda}                   & 0.352                    & 7.20                          & 44.98                  & 0.845            & 0.519           & 0.347              & 8.32                   & 47.06                  & 0.494            & 0.379           \\
		                         & DAC\cite{Guo2025DepthAnyCamera}            & 0.185                    & 15.80                         & 73.81                  & 0.632            & 0.288           & 0.249              & 12.14                  & 66.42                  & 0.394            & 0.264           \\
		                         & DAP\cite{lin2025dap}                       & 0.460                    & 5.23                          & 35.79                  & 1.209            & 0.718           & 0.113              & 39.96                  & 91.03                  & 0.328            & 0.163           \\
		                         & DA360\cite{jiang2025depth}                 & 0.077                    & 59.47                         & 92.59                  & 0.740            & 0.177           & 0.048              & 75.67                  & 98.19                  & 0.273            & 0.059           \\
		                         & UniK3D\cite{piccinelli2025unik3d}          & 0.097                    & 30.30                         & 93.41                  & 0.534            & 0.175           & 0.143              & 26.60                  & 88.40                  & 0.240            & 0.137           \\
		                         & DA$^{2}$\cite{li2025depth}                 & 0.109                    & 42.98                         & 89.89                  & 0.598            & 0.215           & 0.143              & 38.62                  & 89.07                  & 0.285            & 0.158           \\
		                         & VGGT360\cite{wang2025vggt}                 & 0.057                    & 56.35                         & 95.80                  & 0.458            & 0.133           & 0.029              & 81.72                  & 98.66                  & 0.165            & 0.043           \\
		                         & MapAnything360\cite{keetha2025mapanything} & 0.066                    & 49.24                         & 95.28                  & 0.442            & 0.135           & 0.037              & 71.59                  & 98.50                  & 0.151            & 0.048           \\
		                         & $\pi^3$360\cite{wang2025pi}                & 0.059                    & 53.17                         & 95.48                  & 0.466            & 0.142           & 0.037              & 72.61                  & 98.00                  & 0.208            & 0.061           \\
		                         & Ours                                       & \textbf{0.053}           & \textbf{63.34}                & \textbf{96.27}         & \textbf{0.412}   & \textbf{0.114}  & \textbf{0.020}     & \textbf{89.82}         & \textbf{98.96}         & \textbf{0.144}   & \textbf{0.031}  \\
				\midrule
				\multirow{10}*{Median}
		                         & PanDA\cite{cao2025panda}                   & 0.409                    & 5.23                          & 36.95                  & 1.158            & 0.723           & 0.390              & 5.53                   & 38.61                  & 0.809            & 0.568           \\
		                         & DAC\cite{Guo2025DepthAnyCamera}            & 0.183                    & 14.42                         & 73.44                  & 0.647            & 0.299           & 0.213              & 12.59                  & 67.76                  & 0.451            & 0.281           \\
		                         & DAP\cite{lin2025dap}                       & 0.422                    & 5.61                          & 37.12                  & 1.231            & 0.710           & 0.132              & 37.38                  & 90.60                  & 0.353            & 0.181           \\
		                         & DA360\cite{jiang2025depth}                 & 0.123                    & 49.01                         & 91.54                  & 1.002            & 0.227           & 0.054              & 74.09                  & 98.33                  & 0.374            & 0.073           \\
		                         & UniK3D\cite{piccinelli2025unik3d}          & 0.107                    & 26.52                         & 91.99                  & 0.545            & 0.193           & 0.139              & 24.50                  & 87.04                  & 0.266            & 0.156           \\
		                         & DA$^{2}$\cite{li2025depth}                 & 0.126                    & 36.65                         & 89.08                  & 0.622            & 0.239           & 0.150              & 36.84                  & 88.59                  & 0.311            & 0.178           \\
		                         & VGGT360\cite{wang2025vggt}                 & 0.063                    & 50.03                         & 95.61                  & 0.461            & 0.143           & 0.034              & 80.02                  & 98.65                  & 0.168            & 0.047           \\
		                         & MapAnything360\cite{keetha2025mapanything} & 0.069                    & 45.90                         & 95.45                  & 0.456            & 0.144           & 0.040              & 69.73                  & 98.50                  & 0.157            & 0.053           \\
		                         & $\pi^3$360\cite{wang2025pi}                & 0.065                    & 47.68                         & 95.20                  & 0.469            & 0.152           & 0.038              & 70.89                  & 97.98                  & 0.212            & 0.065           \\
		                         & Ours                                       & \textbf{0.055}           & \textbf{59.72}                & \textbf{96.39}         & \textbf{0.418}   & \textbf{0.119}  & \textbf{0.022}     & \textbf{89.02}         & \textbf{98.96}         & \textbf{0.145}   & \textbf{0.033}  \\
		\midrule
		\multirow{4}*{Metric}
		                         & DAC\cite{Guo2025DepthAnyCamera}            & 0.312                    & 6.61                          & 55.40                  & 0.702            & 0.419           & 0.668              & 3.43                   & 36.14                  & 0.571            & 0.501           \\
		                         & UniK3D\cite{piccinelli2025unik3d}          & 0.314                    & 0.96                          & 42.54                  & 0.787            & 0.513           & 0.623              & 0.10                   & 17.01                  & 0.735            & 0.604           \\
		                         & MapAnything360\cite{keetha2025mapanything} & {0.119}                  & {9.00}                        & {91.85}                & {0.497}          & {0.230}         & {0.119}            & {6.35}                 & {97.87}                & {0.219}          & {0.141}         \\
		                         & Ours                                       & \textbf{0.057}           & \textbf{55.61}                & \textbf{96.33}         & \textbf{0.420}   & \textbf{0.122}  & \textbf{0.043}     & \textbf{89.53}         & \textbf{98.93}         & \textbf{0.145}   & \textbf{0.031}  \\

		\bottomrule
	\end{tabular}
\end{table}
\subsection{Monocular Depth Estimation}
We further evaluate Argus on monocular depth estimation for both the real-world and synthetic subsets of Realsee3D.
All methods are evaluated with their largest model weights.
As shown in \cref{tab:mono_depth},
Argus achieves the best overall performance across both subsets,
consistently improving over strong multi-view geometry baselines such as VGGT360 and MapAnything360.
Notably, our gains are most pronounced on the synthetic subset,
where the accurate metric prediction and globally consistent geometry produced by Argus lead to lower AbsRel/RMSE and higher $\delta$ accuracies.
These results indicate that Argus not only recovers reliable camera poses at the scene level,
but also produces high-quality per-view depth that benefits downstream reconstruction.
\subsection{Zero-Shot Performance}
\begin{figure}[tb]
	\centering
	\includegraphics[width=1\linewidth]{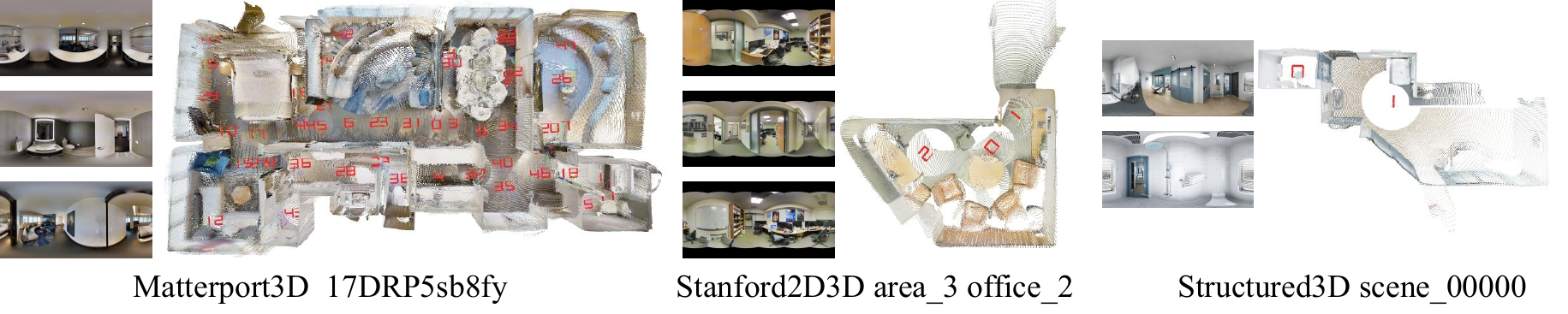}
	\caption{
		\textbf{Zero-Shot Reconstruction on Unseen Datasets.}
	}
	\label{fig:other_pano}
\end{figure}
\begin{table}[tb]
	\tiny
	\caption{\textbf{Zero-Shot Monocular Depth Estimation.}}
	\label{tab:zero_shot_mono_depth}
	\centering
	\begin{tabular}{@{}l*{11}{c}@{}}
		\toprule
		\multirow{2}*{Alignment} & \multirow{2}{*}{Method}           & \multicolumn{5}{c}{Matterport3D\cite{chang2017matterport3d}} & \multicolumn{5}{c}{Stanford2D3D\cite{armeni2017joint}}                                                                                                                                                                           \\
		\cmidrule(lr){3-7}
		\cmidrule(lr){8-12}
		                         &                                   & AbsRel$\downarrow$                                           & $\delta_{1}$$\uparrow$                                 & $\delta_{2}$$\uparrow$ & RMSE$\downarrow$ & MAE$\downarrow$ & AbsRel$\downarrow$ & $\delta_{1}$$\uparrow$ & $\delta_{2}$$\uparrow$ & RMSE$\downarrow$ & MAE$\downarrow$ \\
				\midrule
				\multirow{7}*{IRLS}
		                         & PanDA\cite{cao2025panda}          & 0.372                                                        & 6.00                                                   & 39.33                  & 0.891            & 0.681           & 0.323              & 8.24                   & 48.84                  & 0.778            & 0.533           \\
		                         & DAC\cite{Guo2025DepthAnyCamera}   & 0.202                                                        & 13.18                                                  & 69.89                  & 0.612            & 0.400           & 0.188              & 15.18                  & 74.90                  & 0.592            & 0.323           \\
		                         & DAP\cite{lin2025dap}              & 0.440                                                        & 4.98                                                   & 34.77                  & 1.321            & 0.912           & 0.250              & 12.45                  & 64.14                  & 1.007            & 0.531           \\
		                         & DA360\cite{jiang2025depth}        & 0.106                                                        & \textbf{45.50}                                         & 91.56                  & 1.183            & 0.269           & \textbf{0.064}     & \textbf{60.23}         & 95.98                  & 0.588            & 0.143           \\
		                         & UniK3D\cite{piccinelli2025unik3d} & 0.100                                                        & 27.37                                                  & 92.52                  & \textbf{0.424}   & 0.202           & 0.072              & 42.34                  & \textbf{96.78}         & \textbf{0.347}   & \textbf{0.131}  \\
		                         & DA$^{2}$\cite{li2025depth}        & 0.085                                                        & 38.59                                                  & \textbf{94.09}         & 0.532            & \textbf{0.199}  & 0.071              & 38.39                  & 96.59                  & 0.381            & 0.142           \\
		                         & Ours                              & \textbf{0.082}                                               & 41.77                                                  & 92.54                  & 0.460            & 0.215           & 0.072              & 49.69                  & 95.30                  & 0.427            & 0.157           \\
				\midrule
				\multirow{7}*{Median}
		                         & PanDA\cite{cao2025panda}          & 0.432                                                        & 4.85                                                   & 34.08                  & 1.655            & 0.891           & 0.346              & 6.97                   & 45.29                  & 1.022            & 0.672           \\
		                         & DAC\cite{Guo2025DepthAnyCamera}   & 0.186                                                        & 13.19                                                  & 71.41                  & 0.665            & 0.414           & 0.179              & 14.85                  & 75.49                  & 0.619            & 0.330           \\
		                         & DAP\cite{lin2025dap}              & 0.397                                                        & 5.47                                                   & 37.02                  & 1.362            & 0.901           & 0.249              & 11.70                  & 64.25                  & 1.045            & 0.548           \\
		                         & DA360\cite{jiang2025depth}        & 0.257                                                        & \textbf{41.72}                                         & 93.57                  & 3.300            & 0.539           & 0.101              & \textbf{56.39}         & 95.99                  & 0.770            & 0.166           \\
		                         & UniK3D\cite{piccinelli2025unik3d} & 0.106                                                        & 24.79                                                  & 91.38                  & \textbf{0.440}   & 0.219           & 0.075              & 38.73                  & \textbf{96.39}         & \textbf{0.355}   & \textbf{0.142}  \\
		                         & DA$^{2}$\cite{li2025depth}        & 0.089                                                        & 36.51                                                  & \textbf{93.64}         & 0.549            & \textbf{0.214}  & \textbf{0.073}     & 36.07                  & 96.36                  & 0.391            & 0.151           \\
		                         & Ours                              & \textbf{0.085}                                               & 38.18                                                  & 92.28                  & 0.478            & 0.231           & 0.074              & 44.86                  & 95.06                  & 0.433            & 0.162           \\
		\midrule
		\multirow{3}*{Metric}
		                         & DAC\cite{Guo2025DepthAnyCamera}   & 0.306                                                        & 9.31                                                   & 53.45                  & 0.772            & 0.561           & 0.251              & 10.26                  & 65.47                  & 0.615            & 0.389           \\
		                         & UniK3D\cite{piccinelli2025unik3d} & 0.243                                                        & 3.44                                                   & 63.50                  & \textbf{0.643}   & \textbf{0.447}  & 0.181              & 4.24                   & 85.22                  & 0.499            & 0.331           \\
		                         & Ours                              & \textbf{0.165}                                               & \textbf{14.05}                                         & \textbf{78.66}         & 0.736            & 0.504           & \textbf{0.103}     & \textbf{25.29}         & \textbf{92.04}         & \textbf{0.494}   & \textbf{0.235}  \\

		\bottomrule
	\end{tabular}
\end{table}
We further test the zero-shot generalization of Argus on two mainstream indoor benchmarks,
Matterport3D~\cite{chang2017matterport3d} and Stanford2D3D~\cite{armeni2017joint}.
As summarized in \cref{tab:zero_shot_mono_depth},
Argus delivers strong monocular depth performance on both datasets,
is competitive with specialized panoramic depth baselines,
and holds a notable performance lead in most metric depth evaluation,
demonstrating robust cross-dataset transfer.

To further validate generalization,
we run inference on scenes from Matterport3D~\cite{chang2017matterport3d}, Stanford2D3D~\cite{armeni2017joint}, and Structured3D~\cite{zheng2020structured3d}---none of which are used for training.
As shown in \cref{fig:other_pano}, Argus produces consistent and accurate reconstructions across diverse unseen datasets.
\subsection{Training Data Ablation}
The synthetic-to-real split of Realsee3D is set to 9:1,
an empirically balanced ratio.
Synthetic data boosts model accuracy,
while real data strengthens generalization.
To further dissect their individual contributions,
we conduct ablation experiments training solely on real or synthetic data,
as summarized in \cref{tab:data_ablations}.
\begin{table}[tb]
	\caption{\textbf{Training Data Ablation Results on Realsee3D.} }
	\label{tab:data_ablations}
	\centering
	\begin{tabular}{@{}l*{6}{c}@{}}
		\toprule
		\multirow{2}{*}{Method} & \multicolumn{3}{c}{Real} & \multicolumn{3}{c}{Synthetic}                                                                       \\
		\cmidrule(lr){2-4}
		\cmidrule(lr){5-7}

		                        & AbsRel$\downarrow$       & Acc.$\downarrow$              & ATE$\downarrow$& AbsRel$\downarrow$ & Acc.$\downarrow$ & ATE$\downarrow$\\
		\midrule
		Full                    & 0.050                    & \textbf{0.056}                & \textbf{0.096} & \textbf{0.035}   & \textbf{0.020} & \textbf{0.027} \\
		Real Only               & \textbf{0.047}           & 0.060                         & 0.099          & 0.289            & 0.171          & 0.796          \\
		Synthetic Only          & 0.092                    & 0.156                         & 0.473          & 0.044            & 0.021          & 0.049          \\
		\bottomrule
	\end{tabular}
\end{table}
Experiments verify the complementary effects of real and synthetic data.
Notably, training solely on real data yields slightly better AbsRel than the full mixed dataset.
We attribute this counterintuitive observation to two factors:
the limited volume of real scenes introduces overfitting risks,
and real captures contain sparse depth signals alongside prominent noise and artifacts such as camera calibration inaccuracies.
Thus, we argue that evaluation on the synthetic subset delivers more stable results.
\begin{table}[tb]
	\caption{\textbf{ATE Evaluation of Scalability of Covisibility Module on Real Subset.} }
	\label{tab:covis_scalability}
	\centering
	\begin{tabular}{llll}
		\toprule
		\# Image Number                 & $<$20                  & 20$\sim$30                      & $>$30                  \\
		\midrule
		without reference view learning & 0.092                  & 0.138                           & 0.233                  \\
		with reference view learning    & \textbf{0.082} (-11\%) & \textbf{0.087} (\textbf{-37\%}) & \textbf{0.152} (-35\%) \\
		\bottomrule
	\end{tabular}
\end{table}
\subsection{Scalability of Covisibility Module}
We further analyze the scalability of the Covisibility Module by evaluating ATE on real-world scenes grouped by viewpoint count,
with results presented in \cref{tab:covis_scalability}.
As scene scale grows (from fewer than 20 views to more than 30),
the ATE without reference view learning degrades sharply (0.092$\to$0.233),
indicating that naive first-frame anchoring becomes increasingly fragile with more viewpoints due to accumulated drift.
In contrast, the covisibility-guided reference selection maintains substantially lower error across all scales.
The relative improvement is most pronounced in the 20$\sim$30 range (37\% ATE reduction),
which closely matches the average viewpoint count per scene in our dataset and represents the typical operating regime.
For scenes with more than 30 viewpoints, the absolute gain remains large,
yet the relative improvement is slightly smaller (35\%),
as accumulated drift from distant viewpoints inevitably degrades pose accuracy even with an optimally chosen reference frame.
This demonstrates that our lightweight covisibility module is particularly effective within the typical scene scale and generalizes well to larger environments.
\section{More Visualization}
\subsection{Distribution of Covisibility Scores}
\begin{figure}[tb]
	\centering
	\includegraphics[width=1\linewidth]{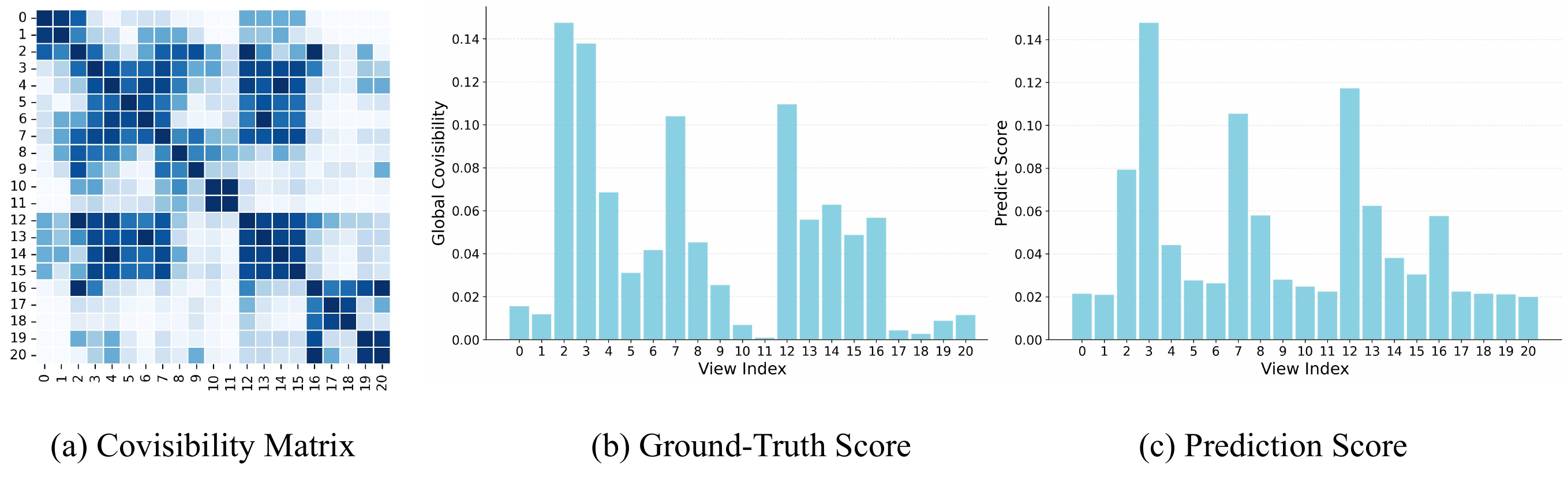}

	\caption{
		\textbf{Visualization of the Covisibility Adjacency Matrix, Ground-Truth and Predicted Global Covisibility Scores.}
	}
	\label{fig:ref}
\end{figure}
\begin{figure}[tb]
	\centering
	\includegraphics[width=1\linewidth]{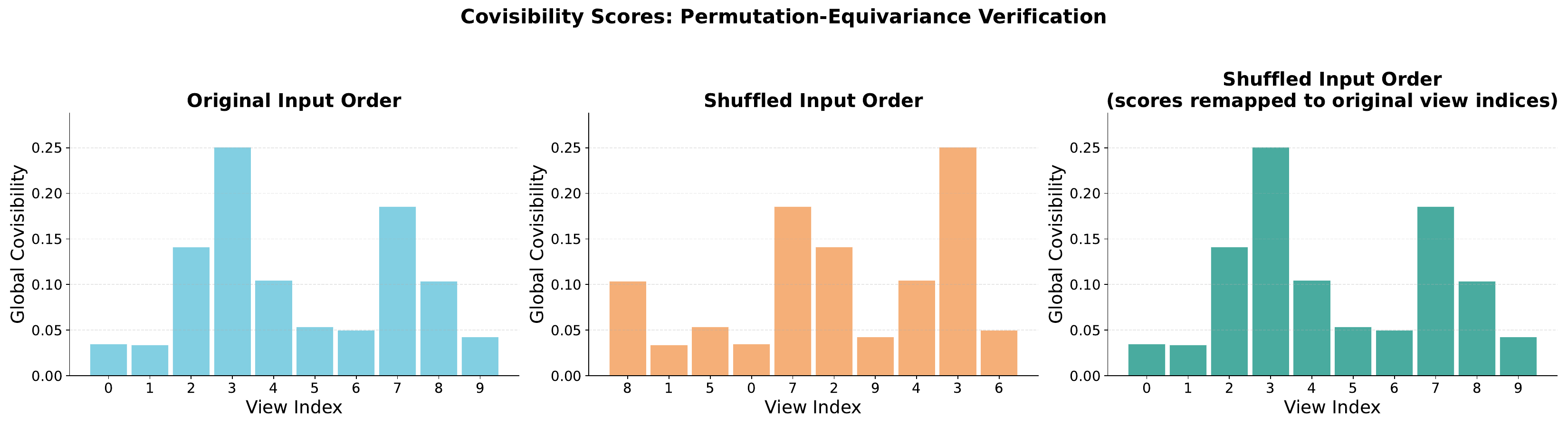}
	\caption{
		\textbf{Verification of Covisibility Score Permutation Equivariance. }
	}
	\label{fig:covisibility_equivariance}
\end{figure}
We take \texttt{scene\_00530} in \cref{fig:ref_vis} as an example to visualize the effectiveness of reference view learning.
In part (a) in \cref{fig:ref}, we visualize the covisibility among all 21 viewpoints in the scene as an adjacency matrix,
where darker colors indicate stronger covisibility.
In (b) and (c), we report the global covisibility scores of each viewpoint,
computed from the ground-truth (GT) and predicted by our network, respectively.
We observe that Argus successfully learns the overall distribution of the GT scores.
Interestingly, the predicted best reference viewpoint is number 3, which corresponds to the second-best choice under the GT ranking.
This suggests that the network does not need to identify the exact optimal view;
selecting a reasonably good reference from unordered inputs is sufficient to substantially improve reconstruction robustness.
Furthermore, our Covisibility Module exhibits permutation equivariance.
As illustrated in \cref{fig:covisibility_equivariance},
randomly shuffling the order of input images barely affects the selection of reference frame.
\section{More Discussions}
\label{sec:discussions}
\subsection{Why Panoramas.}
\begin{figure}[tb]
	\centering
	\includegraphics[width=1\linewidth]{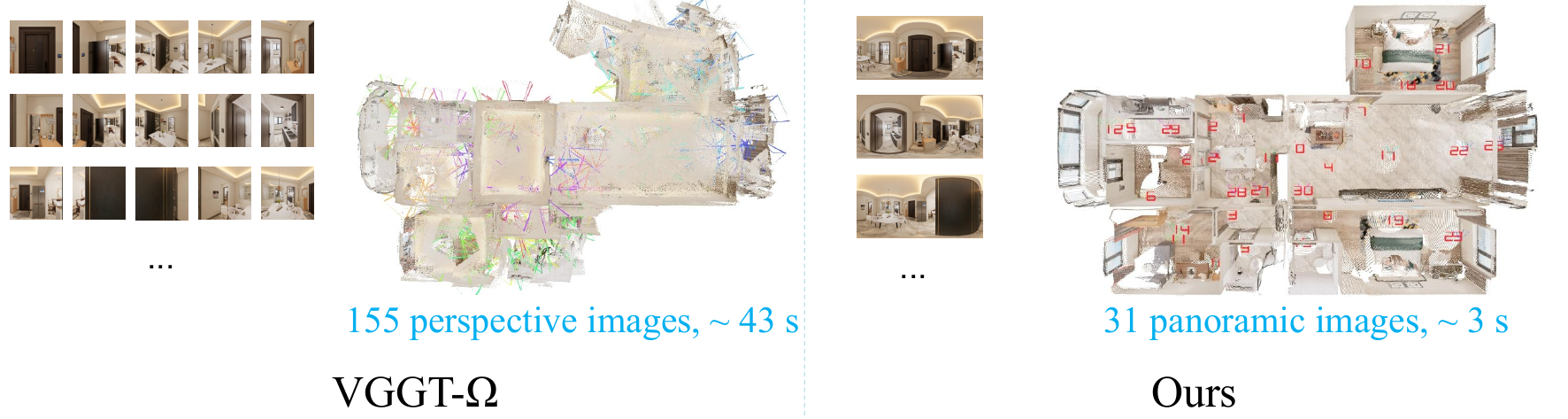}
	\caption{
		\textbf{Perspective and Panoramic 3D Reconstruction.}
		Evaluated on a single RTX 3090 GPU.
	}
	\label{fig:native_comp}
\end{figure}

As illustrated in \cref{fig:native_comp},
panoramic images deliver superior field-of-view coverage for 3D reconstruction compared with perspective images.
We split each panorama into five overlapping perspective views as input to VGGT-$\Omega$~\cite{wang2026vggt},
a recent SOTA perspective-based model.
For identical scenes,
our method yields faster, more complete, and more accurate reconstructions with metric scale.
\subsection{Metric Scale Normalization Factor.}
We set \(s=10\) to normalize the majority of depth values to the interval \([0,1]\).
The depth distribution is visualized in \cref{fig:depth_dist}.
For bounded indoor scenes from the Realsee3D dataset,
our fixed normalization scheme achieves faster convergence and superior accuracy compared to the per-scene distance normalization adopted by MapAnything~\cite{keetha2025mapanything}.
While the per-scene normalization strategy may offer stronger generalization for highly diverse scenes,
we are unable to verify this claim given the limited scale of our current dataset.
\subsection{Model Simplicity versus Overcomplete Supervision.}
There exists a natural trade-off between architectural simplicity and the richness of supervisory signals.
With sufficient training data and model capacity, simpler architectures with fewer prediction heads can achieve strong performance by relying on implicit geometric reasoning.
However, when data and model scales are relatively limited, as in the current Realsee3D dataset and Argus model,
explicitly adding multiple prediction heads to supervise intermediate geometric representations provides complementary gradient pathways that facilitate multi-task learning and improve overall reconstruction accuracy.
Our ablation studies confirm that these additional heads yield consistent gains under this regime.
Although the extra heads increase training-time computation,
they can be disabled at inference time with zero additional overhead,
making overcomplete supervision a cost-effective strategy that trades moderate extra training resources for meaningful accuracy improvements.
\subsection{Cropping Panorama Poles.}
We crop the top and bottom 15\% of each ERP panorama.
On one hand,
the corresponding polar pixels in the real subset of our Realsee3D dataset are incomplete due to limited camera field of view,
these areas are typically masked,
blurred with Gaussian circles,
or filled with reflective artifacts.
On the other hand,
this 30\% pixel region incurs heavy computational overhead yet covers an extremely narrow field of view,
leading to poor computational efficiency.
Missing point clouds from these cropped regions can be easily complemented using adjacent views or simple plane fitting.
\subsection{Limitations}
Due to the scarcity of outdoor datasets,
our model exhibits suboptimal performance in outdoor and aerial scenarios,
and this limitation will be progressively mitigated with the acquisition of additional datasets.
Moreover, current methods fail to realize arbitrary-scale reconstruction due to inherent computational resource limits
for cases with thousands or tens of thousands of panoramic images,
and scale-agnostic reconstruction therefore represents a highly promising research avenue.
One promising approach is to construct a hierarchical system via retrieval-driven pre-grouping,
executing feed-forward reconstruction and result re-aggregation independently in each group.
Our Covisibility Transformer can be naturally extended to support this functionality,
all while maintaining the lightweight,
end-to-end nature of our reconstruction pipeline.
\subsection{Failure Cases}
\begin{figure}[tb]
	\centering
	\includegraphics[width=1\linewidth]{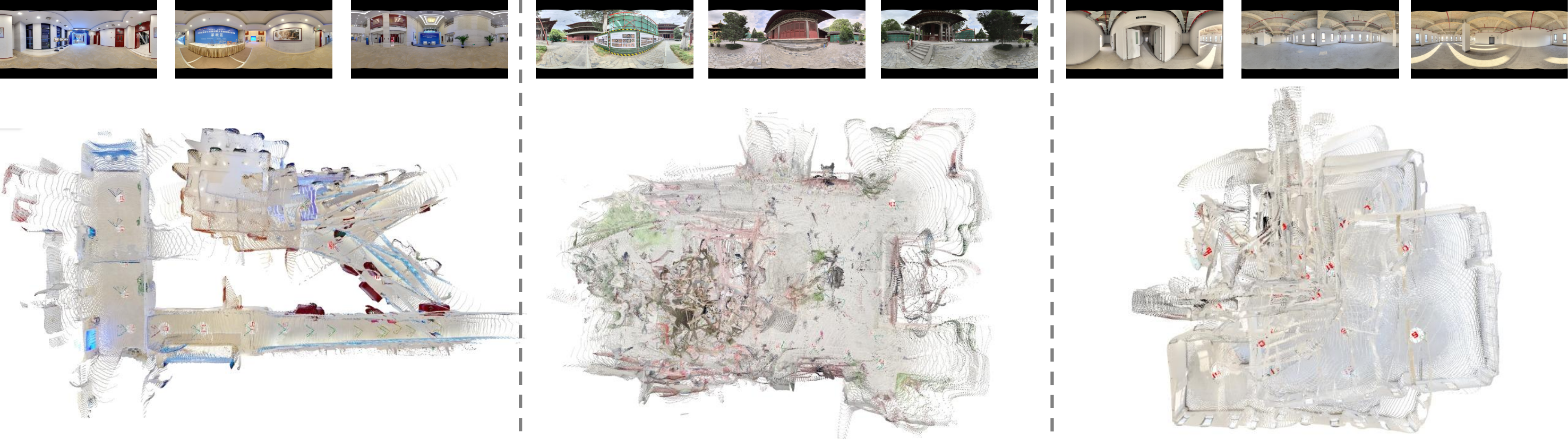}

	\caption{
		\textbf{Failure Cases of Argus.}
	}
	\label{fig:failcase}
\end{figure}
Our method performs stably in typical indoor real estate scenarios.
But as shown in \cref{fig:failcase},
Argus tends to fail in large-scale environments with sparse viewpoints, low texture, and a large number of images,
such as massive conference halls, outdoor parks, and empty factory buildings.
\subsection{Future Work}
Looking ahead,
we will extend Argus to a unified single-model framework that supports multimodal prior inputs,
flexible supervision signals,
and multi-task outputs including 3DGS reconstruction, floor plan generation, semantic or instance segmentation, object detection,
and feature matching,
unlocking its potential for a wider spectrum of real-world applications.
Concurrently, we will keep curating and annotating high-quality panoramic 3D datasets covering diverse scenarios to advance follow-up research in the panoramic 3D vision community.

\end{document}